\documentclass[10pt,twocolumn,letterpaper]{article}

\usepackage{cvpr}              

\usepackage{graphicx}
\usepackage{amsmath}
\usepackage{amssymb}
\usepackage{booktabs}

\usepackage{graphicx}
\usepackage{amsmath}
\usepackage{verbatim}
\usepackage{multirow}
\usepackage{amssymb}
\usepackage{caption}
\usepackage{floatrow}
\usepackage{multirow}
\usepackage{comment}
\usepackage{color}
\usepackage{xcolor}
\usepackage{makecell}
\usepackage{siunitx}
\usepackage{booktabs}
\usepackage{bm}
\usepackage{enumitem}
\usepackage{tabularx}
\usepackage{wrapfig}
\newcommand*{\op}[1]{\operatorname{#1}}

\newcolumntype{I}{!{\vrule width 1.12pt}}

\newcommand{\names}{\textsc{ViT-FL}}
\newcommand{\ViT}{\textsc{ViT}}

\definecolor{gg}{RGB}{15,150,15}
\definecolor{rr}{RGB}{230,45,45}

\usepackage[pagebackref,breaklinks,colorlinks]{hyperref}

\usepackage[capitalize]{cleveref}
\crefname{section}{Sec.}{Secs.}
\Crefname{section}{Section}{Sections}
\Crefname{table}{Table}{Tables}
\crefname{table}{Tab.}{Tabs.}

\def\cvprPaperID{6650} 
\def\confName{CVPR}
\def\confYear{2022}

\begin{document}

\title{Rethinking Architecture Design for Tackling Data Heterogeneity in \\Federated Learning}

\author{Liangqiong Qu$^{1}$\thanks{Equal contribution}  \quad  Yuyin Zhou$^{2*}$  \quad  Paul Pu Liang$^{3*}$  \quad  Yingda Xia$^4$  \quad
Feifei Wang$^1$  \quad
\\   Ehsan Adeli$^1$  \quad  {Li Fei-Fei$^1$  \quad  Daniel Rubin$^1$}
\vspace{1mm}\\
$^1$ Stanford University,$^2$ UC Santa Cruz, $^3$ Carnegie Mellon University, $^4$ Johns Hopkins University \vspace{1mm} \\
{\tt\small {\{liangqiqu, zhouyuyiner, philyingdaxia\}}@gmail.com, pliang@cs.cmu.edu, }  \\
{\tt\small{\{ffwang, eadeli, feifeili, rubin\}@stanford.edu,} } \\
}

\maketitle

\begin{abstract}
   Federated learning is an emerging research paradigm enabling collaborative training of machine learning models among different organizations while keeping data private at each institution. Despite recent progress, there remain fundamental challenges such as the lack of convergence and the potential for catastrophic forgetting across real-world heterogeneous devices. In this paper, we demonstrate that self-attention-based architectures (\emph{e.g.}, Transformers) are more robust to distribution shifts and hence improve federated learning over heterogeneous data. Concretely, we conduct the first rigorous empirical investigation of different neural architectures across a range of federated algorithms, real-world benchmarks, and heterogeneous data splits. Our experiments show that simply replacing convolutional networks with Transformers can greatly reduce catastrophic forgetting of previous devices, accelerate convergence, and reach a better global model, especially when dealing with heterogeneous data. We release our code and pretrained models at \url{https://github.com/Liangqiong/ViT-FL-main} to encourage future exploration in robust architectures as an alternative to current research efforts on the optimization front.
\end{abstract}


\vspace{-3mm}
\section{Introduction}
\vspace{-1mm}
%
%
%
%

\begin{figure}
	\begin{tabular}{c}
    {CWT}\\
	\includegraphics[width=0.9\linewidth]{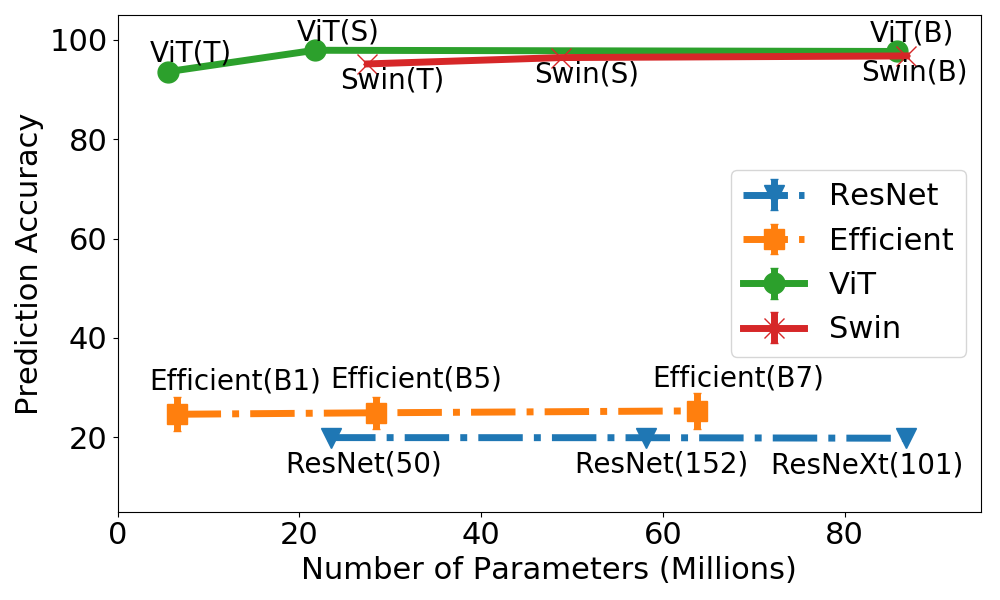} \\
    {FedAVG}\\
    \includegraphics[width=0.9\linewidth]{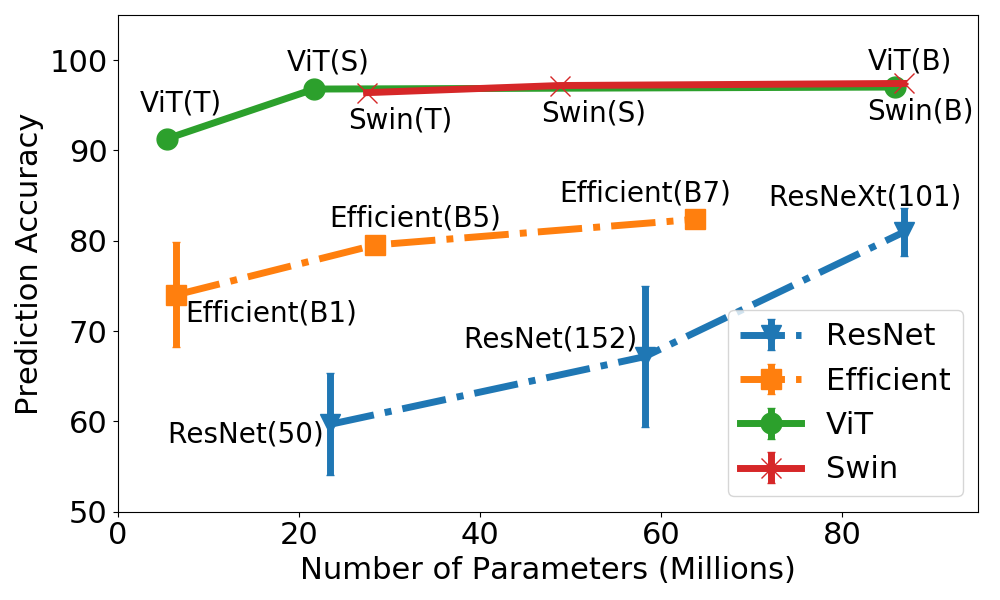} \\
	\end{tabular}
    \vspace{-5mm}
	\caption{Prediction test accuracy on highly heterogeneous data partitions (Split-3) of \textsc{CIFAR-$10$} dataset versus model size\protect\footnotemark[1]. Vision Transformers (ViTs and Swin Transformers) significantly outperform CNNs (ResNets and EfficientNets) on highly heterogeneous data partitions.\vspace{-4mm}}
	\label{fig:acc_vs_parameters}
\end{figure}

Federated Learning (FL) is an emerging research paradigm to train machine learning models on private data distributed over multiple heterogeneous devices~\cite{mcmahan2016communication}. FL keeps data on each device private and aims to train a global model that is updated only via communicated parameters instead of the data itself. Therefore, it provides an opportunity for collaborative machine learning across multiple institutions without risking leakage of private data~\cite{jiang2020federated,li2020federated,rieke2020future}. This has proved especially useful in domains such as healthcare~\cite{brisimi2018federated,chang2018distributed,Guo_2021_CVPR,Liu_2021_CVPR}, learning from mobile devices~\cite{hard2018federated,liang2020think}, smart cities~\cite{jiang2020federated}, and communication networks~\cite{niknam2020federated}, where preserving privacy is crucial. Despite the rich opportunities afforded by FL, there remain fundamental research problems to be tackled before FL can be readily applicable to real-world data distributions. Most current methods that aim to learn a single global model across non-IID devices encounter challenges such as non-guaranteed convergence and model weight divergence for parallel FL methods~\cite{li2018federated,zhao2018federated,Li_2021_CVPR}, and severe catastrophic forgetting problems for serial FL methods \cite{chang2018distributed, gupta2021addressing, sheller2020federated}.

\begin{figure*}
	\centering
	\begin{center}
		\begin{tabular}{c}
	\includegraphics[width=0.9\linewidth]{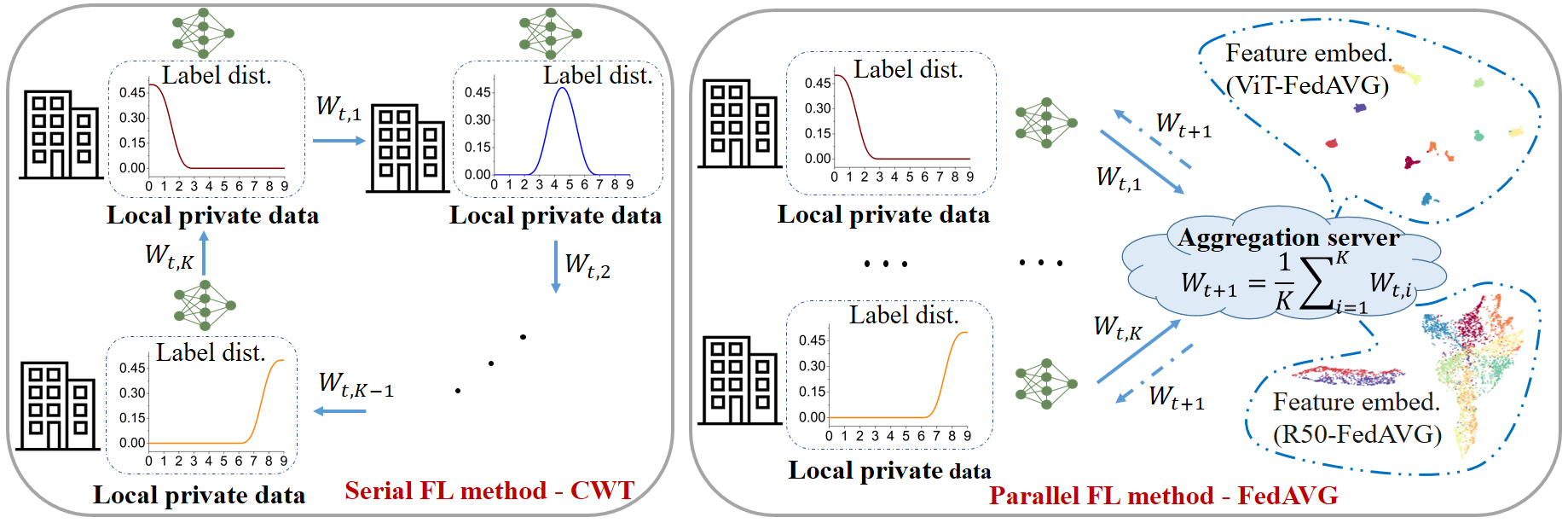}

		\end{tabular}
	\end{center}
\vspace{-6mm}
	\caption{Simplified schematic for a typical serial FL method CWT \cite{chang2018distributed} and a parallel FL method FedAVG \cite{mcmahan2017communication} on non-IID data partitions of CIFAR-$10$~\cite{krizhevsky2009learning} with label distribution skewness. $W_{t,i}$ denotes the model weights during training at round $t$ on client $i$ (total $K$ clients are involved). On the right, we show feature embedding visualizations of ViT(S)-FedAVG and ResNet(50)-FedAVG using UMAP~\cite{mcinnes2018umap}. We find that the features learned by ViT(S)-FedAVG are more clearly separated than those learned by ResNet(50)-FedAVG. Our experiments (section~\ref{sec:results}) support the superiority of \names\ on heterogeneous data and we provide analysis explaining their effectiveness (section~\ref{sec:analysis}).}
	\label{fig:framework_fedavg_CWT}
\end{figure*}

\footnotetext[1]{Mean and standard deviation are calculated across three runs.}

While most research efforts focus on improving the optimization process in FL, our paper aims to provide a new perspective by rethinking the choice of architectures in federated models.
We hypothesize that Transformer architectures~\cite{dosovitskiy2020image,vaswani2017attention} are especially suitable for heterogeneous data distributions due to their surprising robustness to distribution shifts~\cite{bhojanapalli2021understanding}. This property has led to the prevalence of Transformers in self-supervised learning where heterogeneity is manifested via distribution shifts between unlabeled pretraining data and labeled test data~\cite{devlin2018bert}, as well as in multimodal learning over fundamentally heterogeneous input modalities such as image and text~\cite{hu2021transformer,tsai2019multimodal}.
To study this hypothesis, we conduct the first large-scale empirical benchmarking of several neural architectures across a suite of federated algorithms, real-world benchmarks, and heterogeneous data splits.
To represent Transformer networks, we use a standard implementation of Vision Transformers~\cite{dosovitskiy2020image, liu2021swin} on image tasks spanning image classification~\cite{krizhevsky2009learning,liu2015deep} and medical image classification~\cite{KaggleRetina}.

Our results suggest that \names\ (Federated Learning with Vision Transformers) performs especially well in settings with most heterogeneous device splits, with the gap between \names\ and FL with ResNets~\cite{he2016deep} increasing significantly as heterogeneity increases. To understand these results, we find that the main source of improvement lies in the increased robustness of Transformer models to heterogeneous data which reduces catastrophic forgetting of previous devices when trained on substantially different new ones. Together, Transformers converge faster and reach a better global model suitable for most devices. Through comparisons to FL methods designed specifically to combat heterogeneous data, we find that \names\ provides immediate improvements without using training heuristics, additional hyperparameter tuning, or additional training.
Moreover, it is noteworthy that our \names \ is orthogonal to existing optimization based FL methods, and can be easily applied to improve their performance.
To this end, we conclude that Transformers should be regarded as a natural starting point for FL problems in future research.


\vspace{-1mm}
\section{Related Work}
\vspace{-1mm}

\noindent\textbf{Federated Learning.}
Federated learning (FL) aims to train machine learning models on private data across massively distributed devices~\cite{mcmahan2016communication}. To enable effective distributed training across heterogeneous devices, two categories of methods have emerged: (1) parallel FL methods involve training each local client in parallel either synchronously or asynchronously (such as the classic FedAVG~\cite{mcmahan2016communication}), whereas (2) serial methods train each client in a serial and cyclical way (such as Cyclic Weight Transfer (CWT)~\cite{chang2018distributed} and Split learning~\cite{vepakomma2018split}). A schematic description of FedAVG~\cite{mcmahan2016communication} and CWT~\cite{chang2018distributed} is illustrated in Figure~\ref{fig:framework_fedavg_CWT}. At its core, FL presents a challenge of data heterogeneity in the distributions of training data across clients, which causes non-guaranteed convergence and model weight divergence for parallel FL methods ~\cite{hsieh2020non,li2018federated,zhao2018federated,zhang2021splitavg}, and severe catastrophic forgetting problem for serial FL methods~\cite{chang2018distributed, gupta2021addressing, sheller2020federated}.

Among recent developments to the classic FedAVG algorithm~\cite{mcmahan2016communication} have included using server momentum (FedAVGM) to mitigate per-client distribution shift and imbalance~\cite{hsu2019measuring}, globally sharing small subsets of data among all users (FedAVG-Share)~\cite{zhao2018federated}, using a proximal term to the local objective (FedProx) to reduce potential weight divergence~\cite{li2018federated}, or using other optimization heuristics such as collaborative replay~\cite{qu2022handling}, unsupervised contrastive learning~\cite{Zhuang_2021_ICCV}, matching feature layers of user models~\cite{wang2020federated,Zhang_2021_ICCV}, or model distillation~\cite{Gong_2021_ICCV,zhang2021splitavg} to handle heterogeneity.



Concurrently, several recent efforts aim to alleviate catastrophic forgetting in continual and serial learning: constraining the updates on weights that are important to previously seen tasks or clients (elastic weight consolidation (EWC)~\cite{kirkpatrick2017overcoming}), applying Deep Generative Replay to mimic data from previous clients or tasks \cite{shin2017continual,qu2021handling}, and applying cyclically weighted objectives to mitigate performance loss across label distribution skewness~\cite{balachandar2020accounting}, among others.
However, all of these approaches mainly focus on improving the optimization algorithm without studying the potential in architecture design to improve robustness to distribution shifts in data. In our work, we show that simple choices in architecture actually make a big difference and should be an active area of study in parallel to the optimization methods that have been the main focus of current work.

\noindent\textbf{Transformers.}
The Transformer architecture was first proposed for sequence-to-sequence machine translation~\cite{vaswani2017attention} and has subsequently established state-of-the-art performance across many NLP tasks, especially when trained in a self-supervised paradigm~\cite{devlin2018bert}.
Recently, Transformers have also been found to be broadly applicable to tasks involving images and video.
For instance, Parmar \emph{et al.}~\cite{parmar2018image} applied self-attention to local neighborhoods of an image while the Vision Transformer (ViT)~\cite{dosovitskiy2020image} achieved state-of-the-art on ImageNet classification by directly applying Transformers with global self-attention to full-sized images.

Its intriguing performance boosts relative to classical architectures for language (\emph{i.e.}, LSTMs~\cite{HochSchm97}) and vision (\emph{i.e.}, CNNs~\cite{he2016deep,lecun1998gradient}) have inspired recent interest towards understanding the reasons behind their effectiveness. Among several particularly relevant findings are that ViTs are highly robust to severe occlusions, perturbations, domain shifts~\cite{bhojanapalli2021understanding,naseer2021intriguing}, as well as synthetic and natural adversarial examples~\cite{mahmood2021robustness,paul2021vision}. In addition, recent studies have suggested that Transformers are also suitable for heterogeneous and multimodal data~\cite{hu2021transformer,lu2021pretrained,tsai2019multimodal}. Inspired by these findings, we hypothesize that ViTs will also be highly effective in adapting to data heterogeneity in FL, and provide detailed empirical analysis to test this hypothesis.



\vspace{-1mm}
\section{Transformers in Federated Learning}
\vspace{-1mm}

In this section, we present background on Transformer architectures and federated learning methods.

\vspace{-1mm}
\subsection{Vision Architectures}
\vspace{-1mm}

\noindent\textbf{CNN.}
For convolution-based architectures, we use the ResNet~\cite{he2016deep} model family (ResNet-50, ResNet-152, and ResNeXt-101 (32x8d)) and EfficientNet~\cite{tan2019efficientnet} model family (EfficientNet-B1, EfficientNet-B5, and EfficientNet-B7), which contains a sequence of convolution, ReLU, pooling, and batch normalization layers. ResNet and EfficientNet are among the most popular architectures for image classification and have been the standard architecture used in FL on image data~\cite{arivazhagan2019federated,lin2020ensemble}.

\noindent\textbf{Transformers.} As a comparison, we employ Vision Transformers (ViT(S), ViT(T), ViT(B))~\cite{dosovitskiy2020image} model family  and Swin Transformer model family (Swin(T), Swin(S), and Swin(B)) ~\cite{liu2021swin}, which do not use conventional convolution layers. Instead, the image features are extracted with image sequentialization and patch embedding strategies. 
See Figure~\ref{fig:acc_vs_parameters} for the number of parameters for each model.

\vspace{-1mm}
\subsection{Federated Learning Methods}
\vspace{-1mm}

We apply one of the most popular parallel methods (FedAVG~\cite{mcmahan2016communication}) and serial methods (CWT~\cite{chang2018distributed}) as training algorithms (see schematic descriptions in Figure~\ref{fig:framework_fedavg_CWT}).


\noindent\textbf{Federated Averaging.}
FedAVG combines local stochastic gradient descent (SGD) on each client with iterative model averaging~\cite{mcmahan2016communication}. Specifically, a fraction of local clients are randomly sampled in each communication round, and the server sends the current global model to each of these clients. Each selected client then performs $E$ epochs of local SGD on its local training data and sends the local gradients back to the central server for aggregation synchronously. The server then applies the averaged gradients to update its global model, and the process repeats.

\noindent\textbf{Cyclic Weight Transfer.}
In contrast to FedAVG where each local client is trained in a synchronous and parallel way, the local clients in CWT are trained in a serial and cyclic manner. In each round of training, CWT trains a global model on one local client with its local data for a number of epochs $E$, and then transfers this global model to the next client for training, until all local clients have been trained on once~\cite{chang2018distributed}. The training process then cycles through the clients repeatedly until the model converges or a predefined number of communication rounds is reached.

\begin{figure*}[t]
	\centering
	\begin{center}
		\begin{tabular}{cc}
\vspace{1mm}
      { \scriptsize{(a) CWT} } &  {\scriptsize{(b) FedAVG}} \\
        \includegraphics[width=0.425\linewidth]{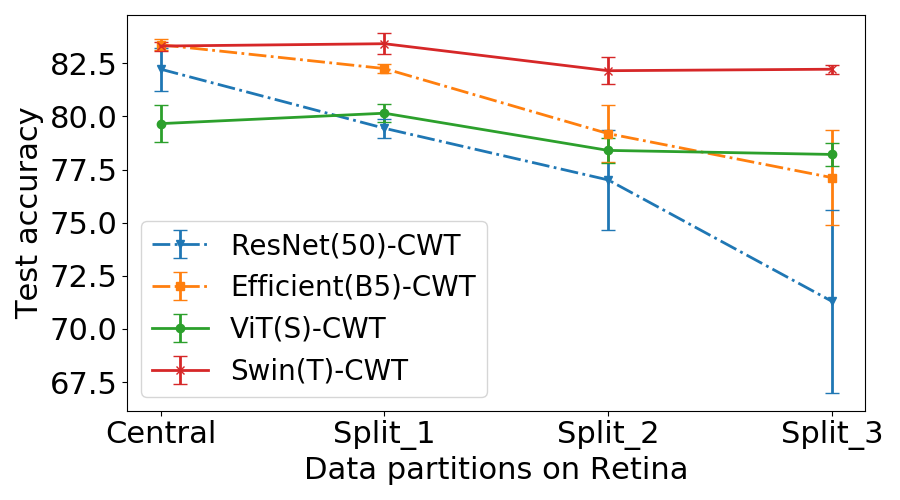} &
        \includegraphics[width=0.425\linewidth]{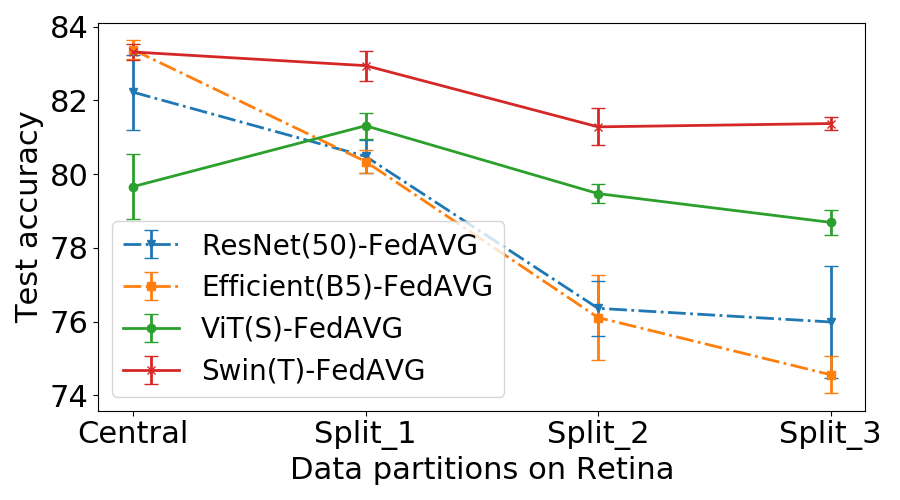}\\
         \includegraphics[width=0.425\linewidth]{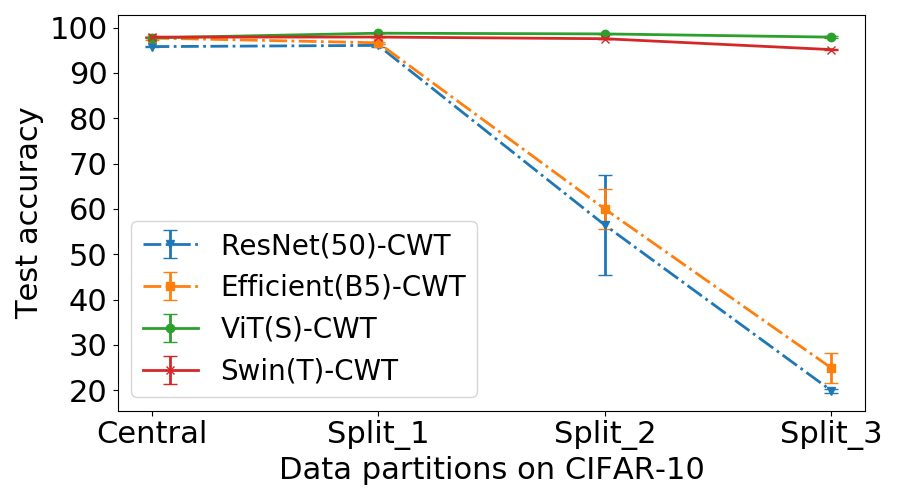} &
        \includegraphics[width=0.425\linewidth]{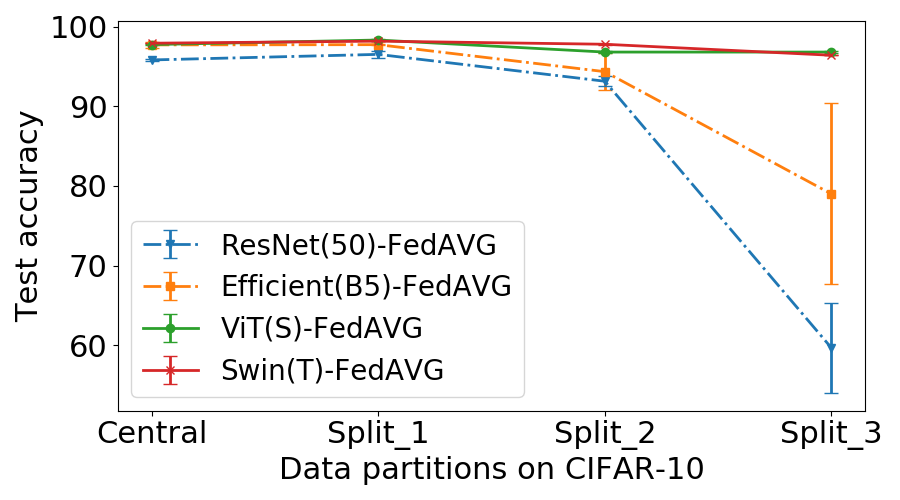}\\
        \end{tabular}
	\end{center}
	\vspace{-5mm}
	\caption{Prediction accuracy (\%) of both CWT and FedAVG with CNNs and Transformers as baseline networks on Retina dataset (first row) and \textsc{CIFAR-$10$} dataset (second row), respectively. Vision Transformers (both ViT and Swin) show consistently strong performance especially in non-IID data partitions.}
    \vspace{-3mm}
    \label{fig:fig_acc_on_Cifar10_Retina}
\end{figure*}

\vspace{-1mm}
\section{Experiments}
\vspace{-1mm}

Our experiments are designed to answer the following research questions that are of importance to practical deployment of FL methods, while also aiding our understanding of (vision) Transformer architectures.
\begin{itemize}[leftmargin=*]
    \setlength\itemsep{0.1em}
    \item Are Transformers able to learn a better global model in FL settings as compared to CNNs which have been the de-facto approach on FL tasks (section~\ref{sec:results})?
    \item Are Transformers especially capable of handling heterogeneous data partitions (section~\ref{sec:generalize})?
    \item Do Transformers reduce communication costs as compared to CNNs (section~\ref{sec:converge})?
    \item Can Transformers be applied to further improve existing optimization-based FL methods (section ~\ref{sec:applied_to_FLs})?
    \item What are practical tips helpful for practitioners to deploy Transformers in FL (section~\ref{sec:tips})?
\end{itemize}

\vspace{-1mm}
\subsection{Experimental Setup}
\vspace{-1mm}

Following~\cite{chang2018distributed,hsieh2020non}, we evaluate FL on the Kaggle Diabetic Retinopathy competition dataset (denoted as Retina)~\cite{KaggleRetina}, CIFAR-$10$ dataset~\cite{krizhevsky2009learning} with simulated data partitions, and a real-world CelebA dataset~\cite{liu2015deep} in our study.

\textbf{Retina and CIFAR-$10$:} We binarize the labels in the Retina dataset to Healthy (positive) and Diseased (negative), randomly selecting $6,000$ balanced images for training, $3,000$ images as the global validation dataset, and $3,000$ images as the global testing dataset following~\cite{chang2018distributed}. We use the original test set in CIFAR-$10$ as the global test dataset, set aside $5,000$ images from the original training dataset as the global validation dataset, and use the remaining $45,000$ images as the training dataset. We simulate three sets of data partitions: one IID-data partition, and two non-IID data partitions with label distribution skew. Each data partition in Retina and CIFAR-$10$ contains $4$ and $5$ simulated clients, respectively. We use the mean Kolmogorov-Smirnov (KS) statistic between every two clients to measure the degree of label distribution skewness. $\textrm{KS}=0$ indicates IID data partitions, while $\textrm{KS}=1$ results in an extremely non-IID data partition, where each client holds totally different label distributions (see Appendix~\ref{appendix:partition} for detailed pre-processing and data partitions).

\textbf{CelebA} is a large-scale face attributes dataset with more than $200$K celebrity images. We use the federated version of CelebA provided by the LEAF benchmark~\cite{caldas2018leaf} which partitions into devices based on identity. Following~\cite{caldas2018leaf}, we test on the binary classification task (presence of smile) and drop clients with larger than $8$ samples to increase task difficulty. This results in a total of $227$ clients each with an average of $5.34\pm1.11$ samples and a total of $1213$ samples.

We use linear learning rate warm-up and decay scheduler for \names. The learning rate scheduler for FL with CNNs is selected from linear warm-up and decay or step decay. Gradient clipping (at global norm 1) is applied to stabilize training. We set the local training epoch $E$ in all FL methods to $1$ (unless otherwise stated), and the total communication rounds to $100$ for Retina and CIFAR-$10$, and $30$ for CelebA. For fair comparison, all models used in this paper are pretrained on ImageNet-1K~\cite{deng2009imagenet}. More implementation details are in Appendix~\ref{appendix:hyperparameter}.

\textbf{Compute:} All experiments were conducted on either a TITAN V GPU or Tesla V100 GPU.

\begin{figure*}[t]
	\centering
	\begin{center}
		\begin{tabular}{ccc}
	\includegraphics[width=0.27\linewidth]{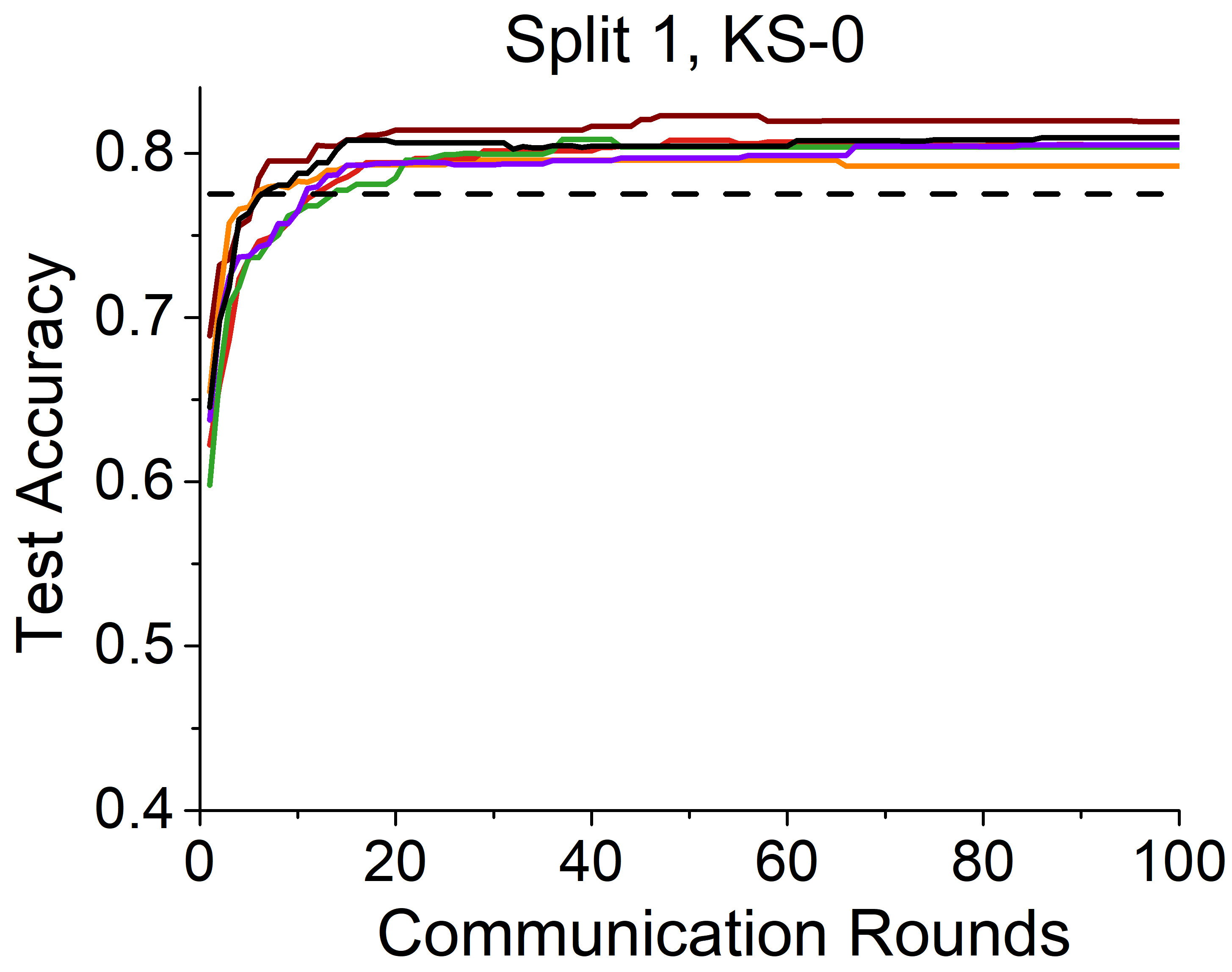}& \hspace{-3mm}
	\includegraphics[width=0.27\linewidth]{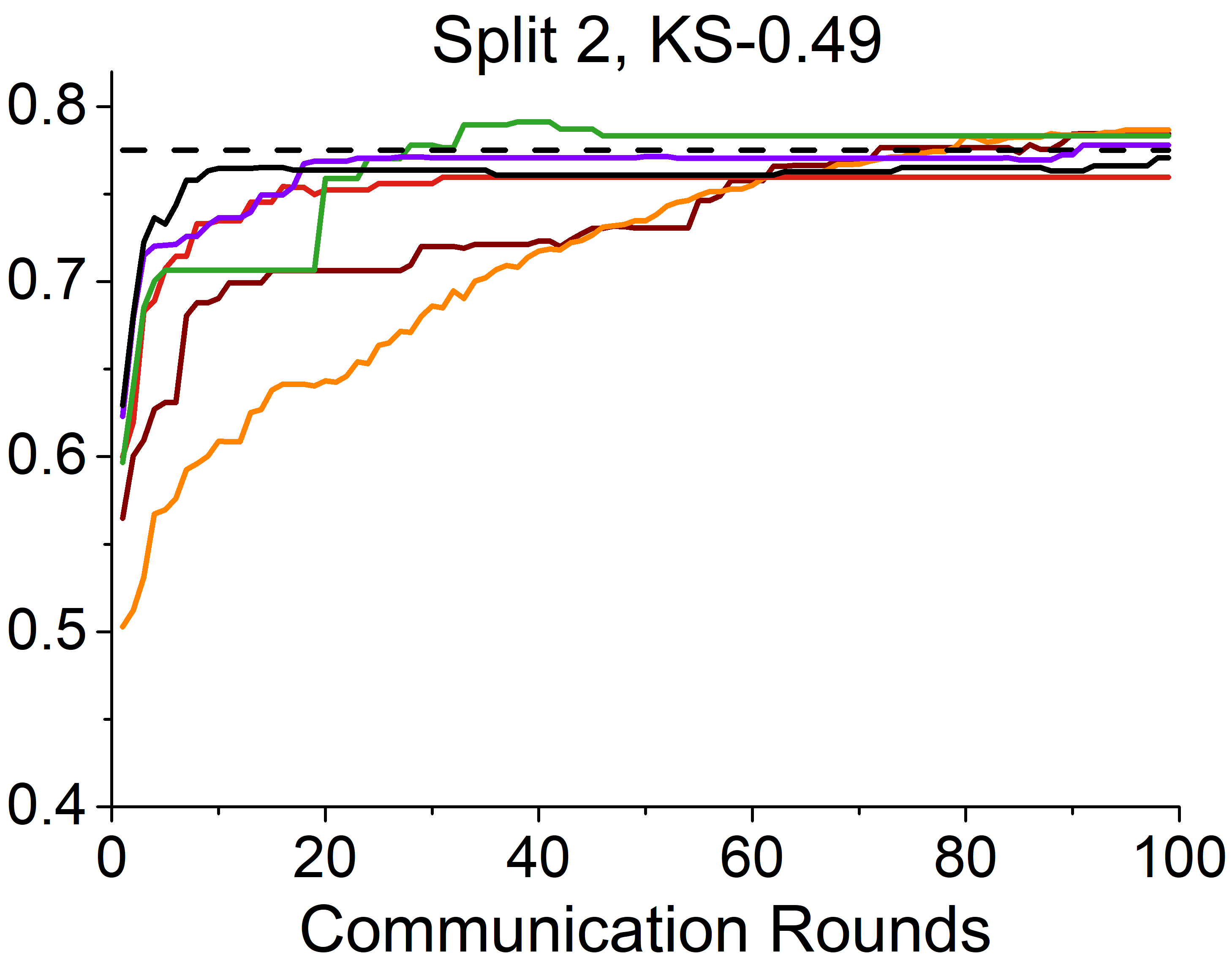}&  \hspace{-3mm}
	\includegraphics[width=0.405\linewidth]{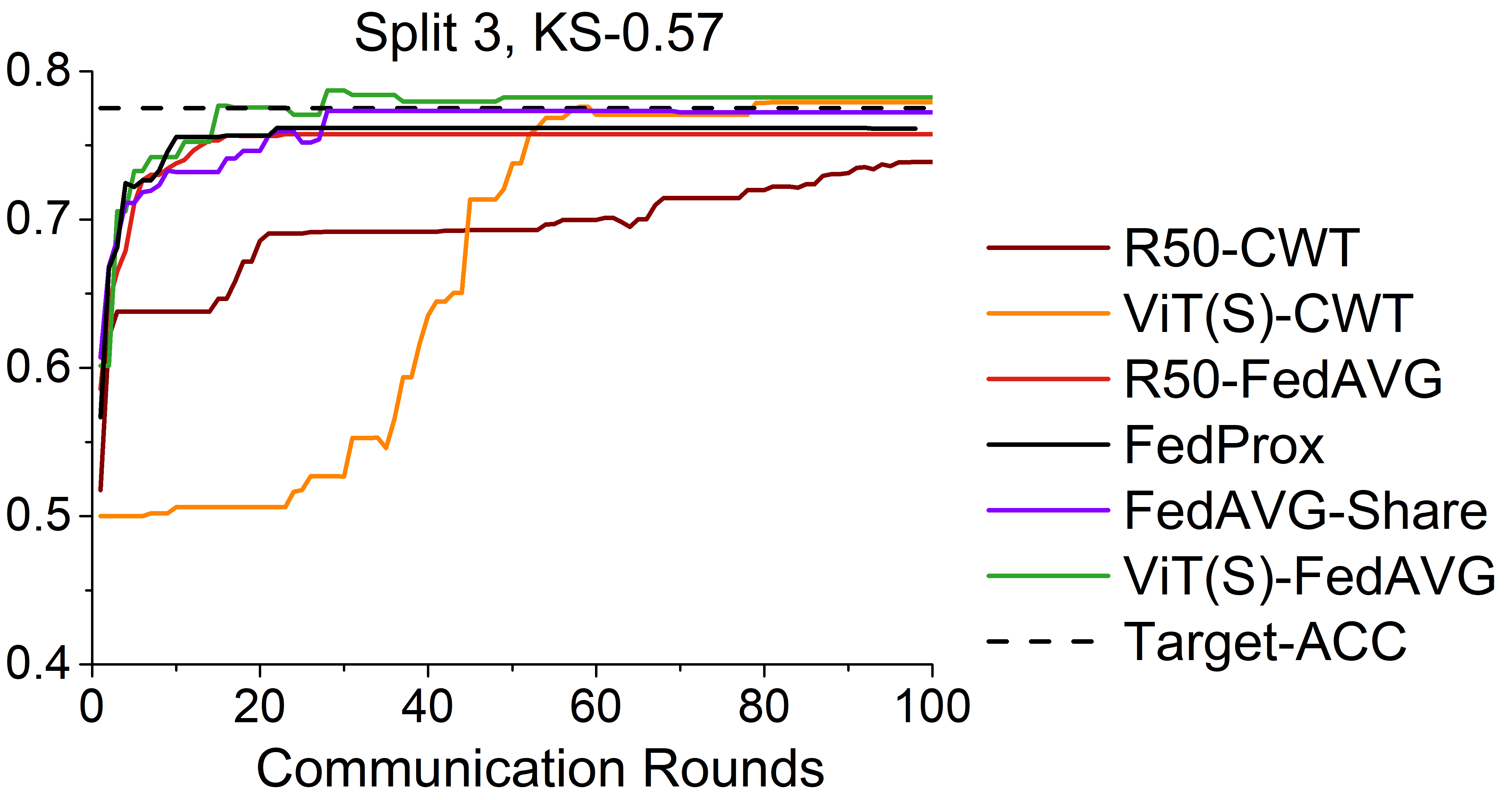} \\
	\includegraphics[width=0.273\linewidth]{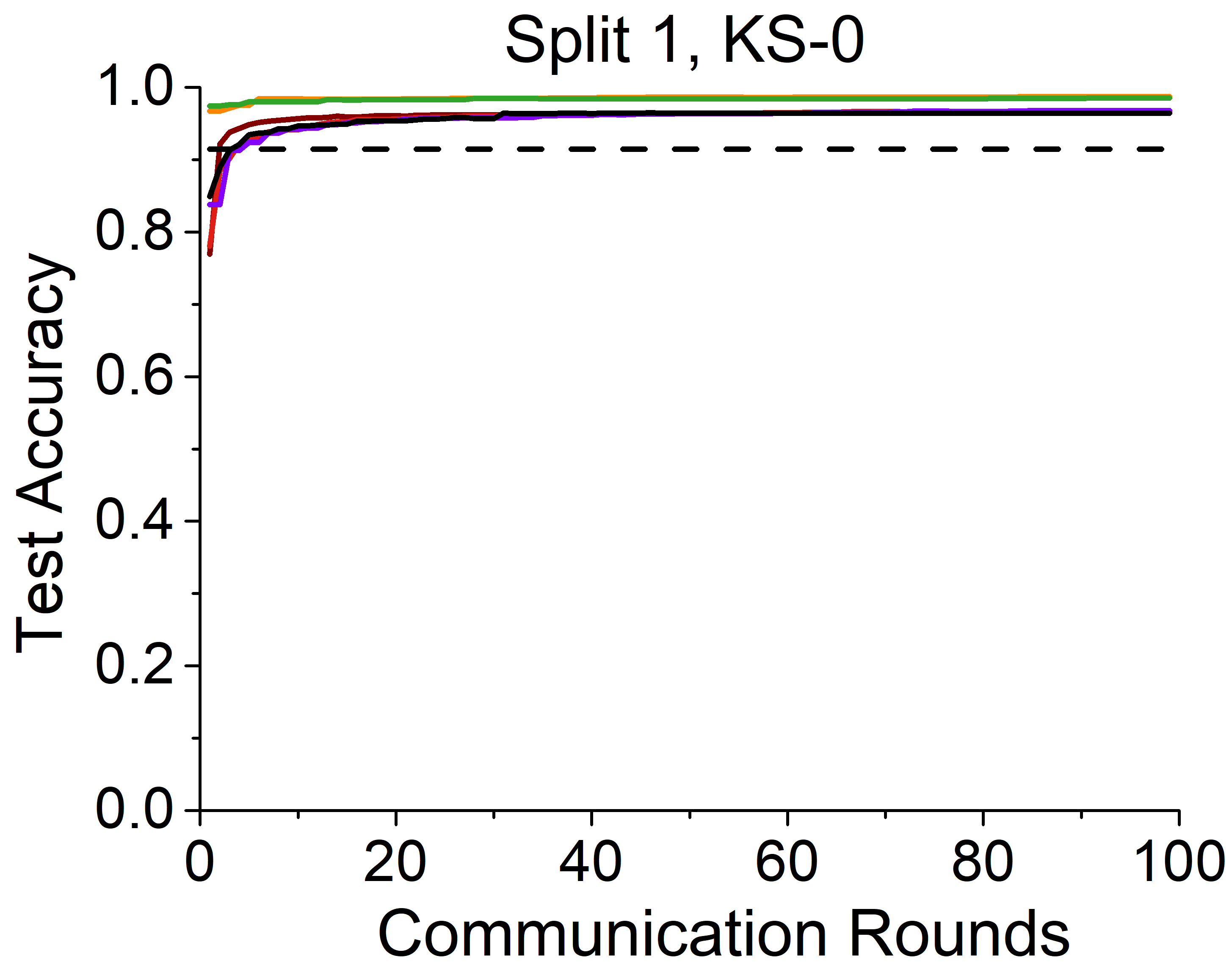}& \hspace{-3mm}
	\includegraphics[width=0.268\linewidth]{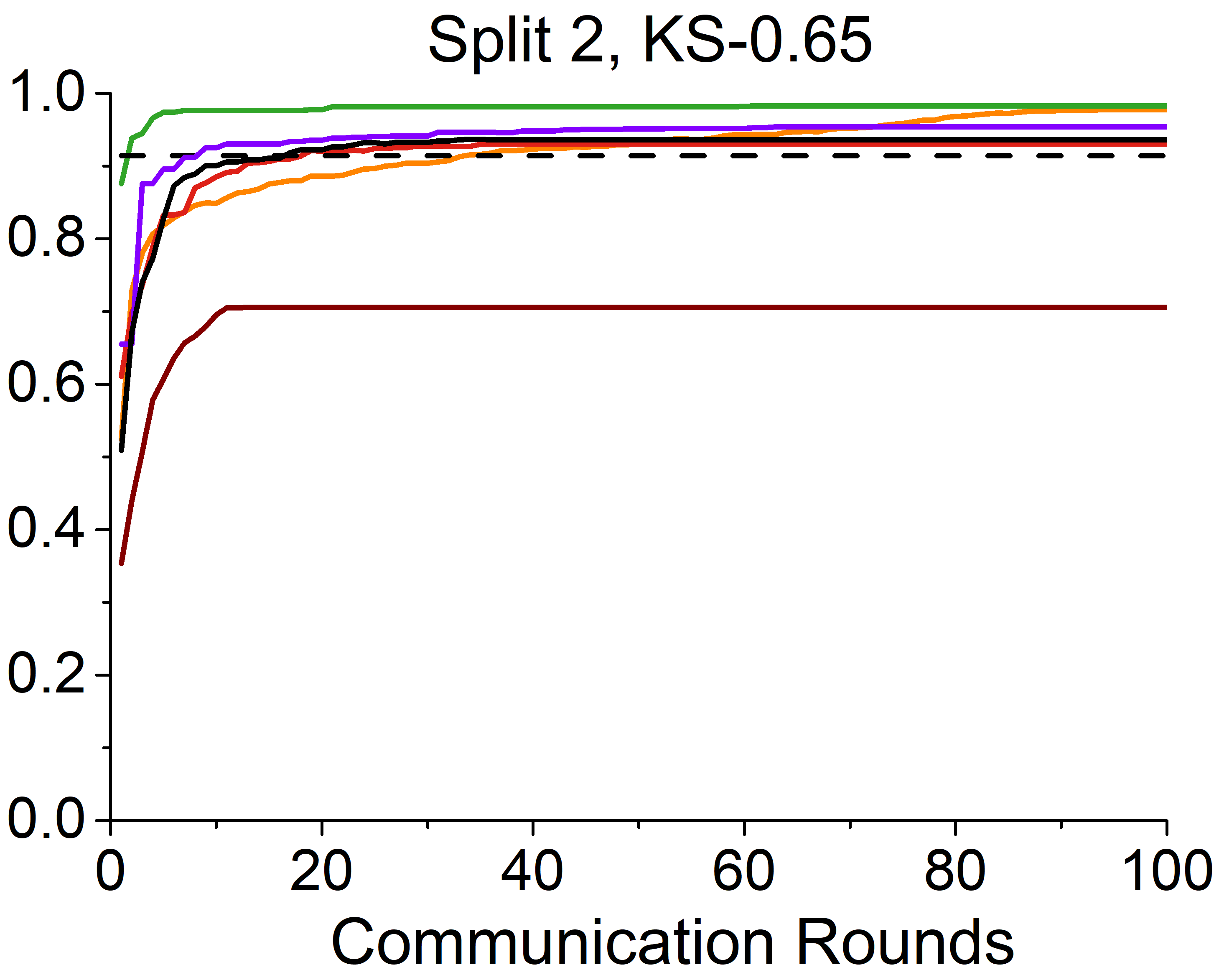}&  \hspace{-3mm}
	\includegraphics[width=0.405\linewidth]{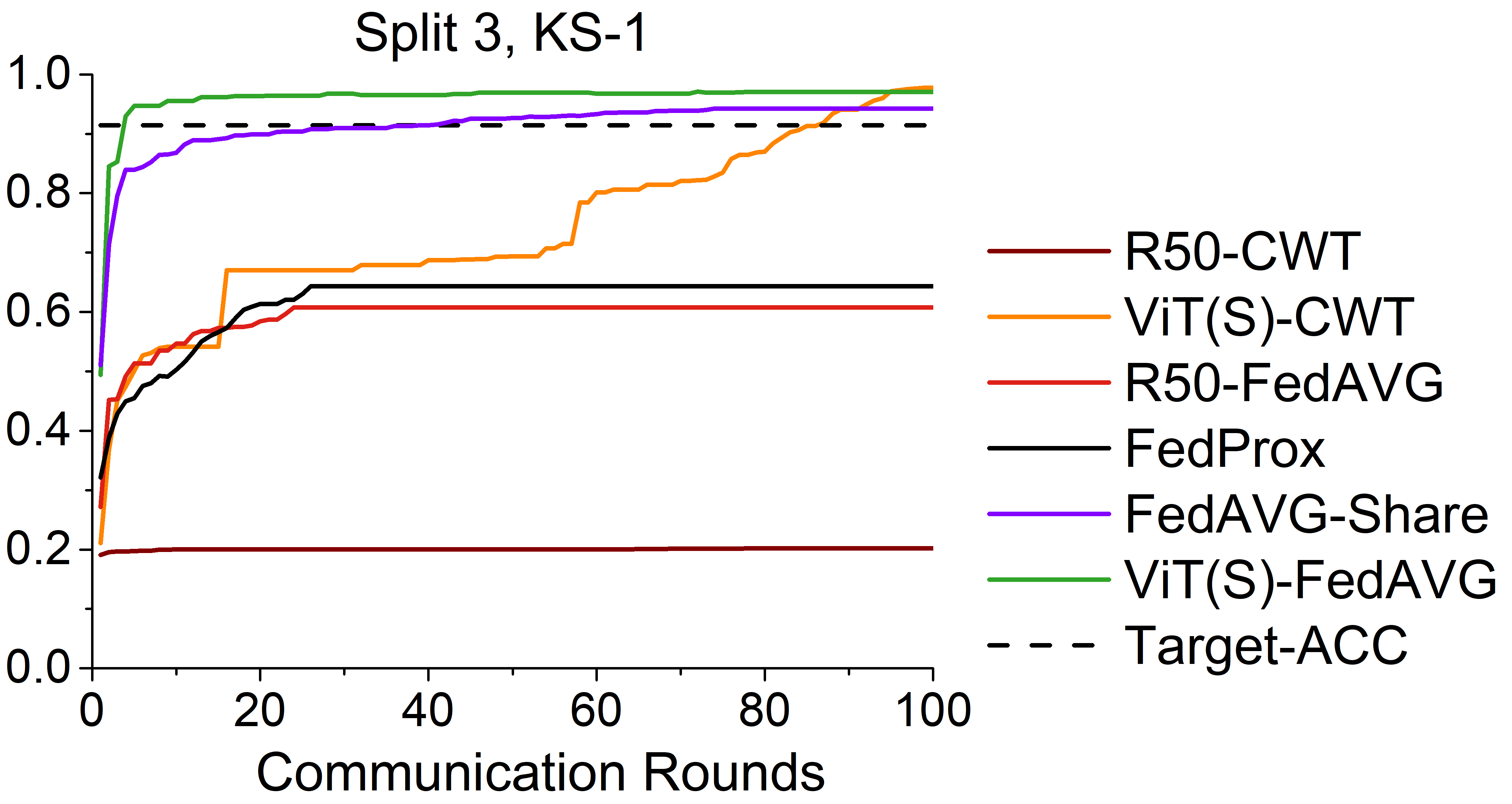} \\
		\end{tabular}
	\end{center}
    \vspace{-6.5mm}
	\caption{Test set accuracy versus communication rounds on Retina dataset (first row) and CIFAR-$10$ dataset (second row) with different data partitions. The black dashed line shows the target performance (Target-ACC) used in Table~\ref{table:communication_costs}. Vision Transformers converge faster with fewer communication rounds, which make them especially suitable for communication-efficient FL.}
	\label{fig:Communication_ViT_ACC}
    \vspace{-1mm}
\end{figure*}

\vspace{-1mm}
\subsection{Results}
\label{sec:results}
\vspace{-1mm}

\textbf{Comparison of FL with different neural architectures and (ideal) centralized training:} Both CWT and FedAVG achieve comparable results to a model trained on centrally hosted data (denoted as Central) on the IID setting no matter which architecture is applied (Figure~\ref{fig:fig_acc_on_Cifar10_Retina}). However, we observe a significant reduction in test accuracy for CNNs on heterogeneous data partitions for both CWT and FedAVG, especially on extremely heterogeneous data partitions (Split 3, KS-1 of CIFAR-$10$) (Figure~\ref{fig:fig_acc_on_Cifar10_Retina} and Figure~\ref{fig:acc_vs_parameters}). By simply replacing CNNs with ViTs, both CWT and FedAVG successfully retain model accuracy even in highly heterogeneous non-IID settings. ViT(S)-CWT and ViT(S)-FedAVG improve the test accuracy relative to ResNet(50)-CWT and ResNet(50)-FedAVG by $77.70\%$ and $37.34\%$ on the highly heterogeneous Split-3, KS-1 of CIFAR-$10$ dataset. Therefore, \ViT\ is particularly suitable for heterogeneous data.

\textbf{Comparison with existing FL methods:} We also compare \names\ to two state-of-the-art optimization based FL methods: FedProx~\cite{li2018federated}, and FedAVG-Share~\cite{zhao2018federated} on both Retina and CIFAR-$10$. We use ResNet(50) as the backbone network for the other compared methods, and ViT(S) for our methods. We tune the best parameters (penalty constant $\mu$ in the proximal term of FedProx) on Split-2 dataset with grid search, and apply the same parameters to all the remaining data partitions.
We allow each client to share $5\%$ percentage of their data among each other for FedAVG-Share. As shown in Figure~\ref{fig:Communication_ViT_ACC}, \names\ outperforms all the other FL methods in non-IID data partitions, especially on the highly heterogeneous non-IID settings.  FedProx~\cite{li2018federated} suffers severe performance drops on highly heterogeneous data partitions despite carefully tuned optimization parameters. Similarly, FedAVG-Share also suffers from performance drops on highly heterogeneous data partition Split-3 even when $5\%$ percentage of the local data is shared among all clients ($94.4\%$ of Split-3 on CIFAR-$10$ dataset compared to $97\%$ on Split-1). We conclude that simply using Transformers outperforms several recent methods designed for FL, which often require careful tuning of optimization parameters. Please note that the usage of \ViT s is orthogonal to the existing optimization methods, and a combination of both can yield
stronger performance (see details in Section~\ref{sec:applied_to_FLs}).

\begin{table*}
  \renewcommand{\arraystretch}{1}
  \centering
  \footnotesize
 \caption{Prediction accuracy ($\%$) on CelebA dataset. Vision Transformers show superior performance to their ResNet(50) (R50 in Table) counterparts, and also outperform the optimization based FL methods (FedProx and FedAVG-Share) with ResNet(50) as backbone network.}
   \label{table:celebA_results}
   \begin{tabular}{c|cc|cccc}
    \Xhline{3\arrayrulewidth}
    \centering

& R50-CWT & ViT(S)-CWT    &  R50-FedAVG & R50-FedProx & R50-FedAVG-Share & ViT(S)-FedAVG \\
    \Xhline{0.5\arrayrulewidth}

CelebA & $85.35 \pm 8.27$ & $\textbf{88.09} \pm \textbf{5.15}$ &  $84.08 \pm 9.65$ & $84.27 \pm 9.74$ & $85.46 \pm 3.75$ & $\textbf{86.63} \pm \textbf{7.12}$  \\
    \Xhline{3\arrayrulewidth}
  \end{tabular}
\end{table*}

\begin{figure*}
	\centering
	\begin{center}
		\begin{tabular}{cc}
	\includegraphics[width=0.385\linewidth]{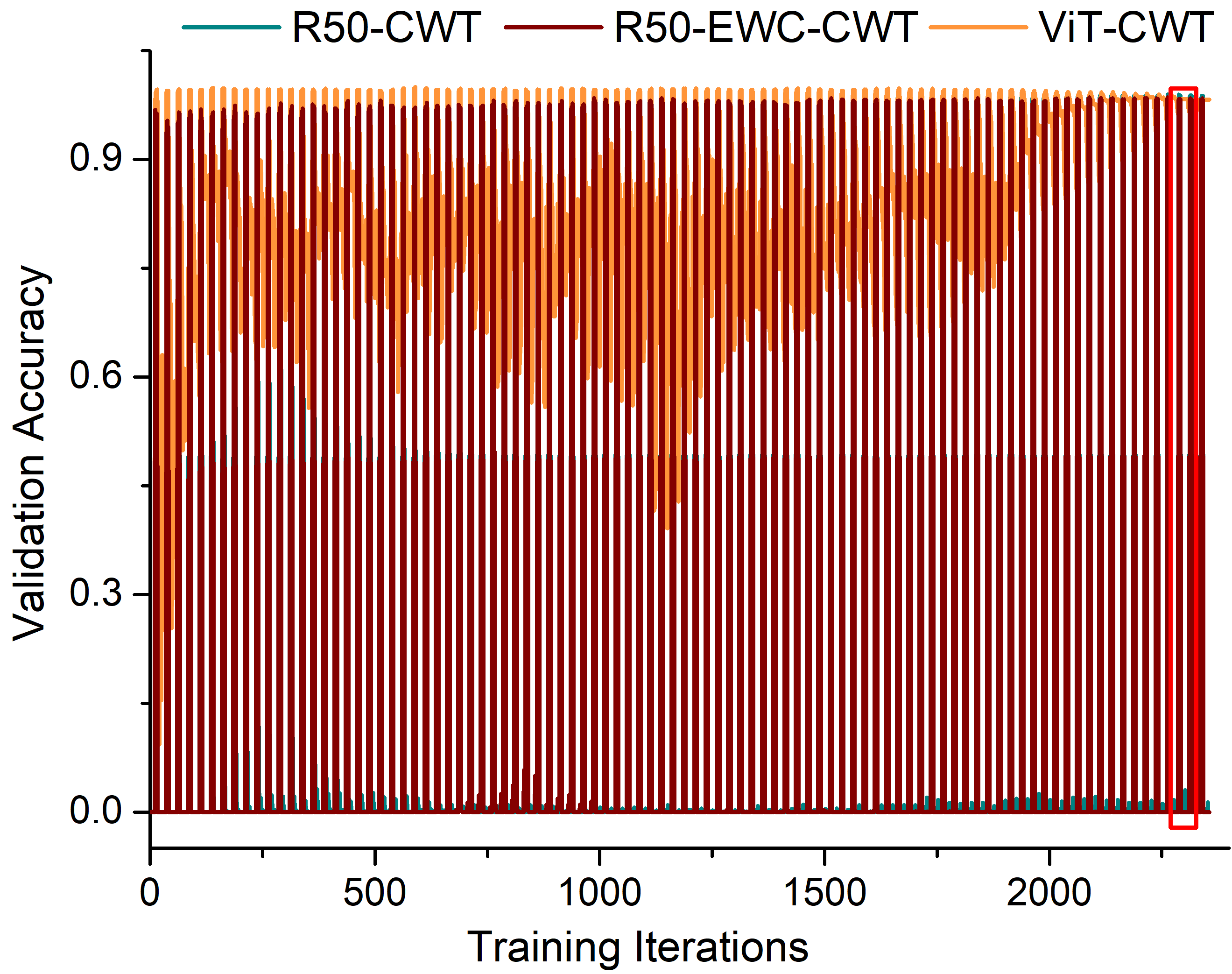}& \hspace{-0mm}
	\includegraphics[width=0.385\linewidth]{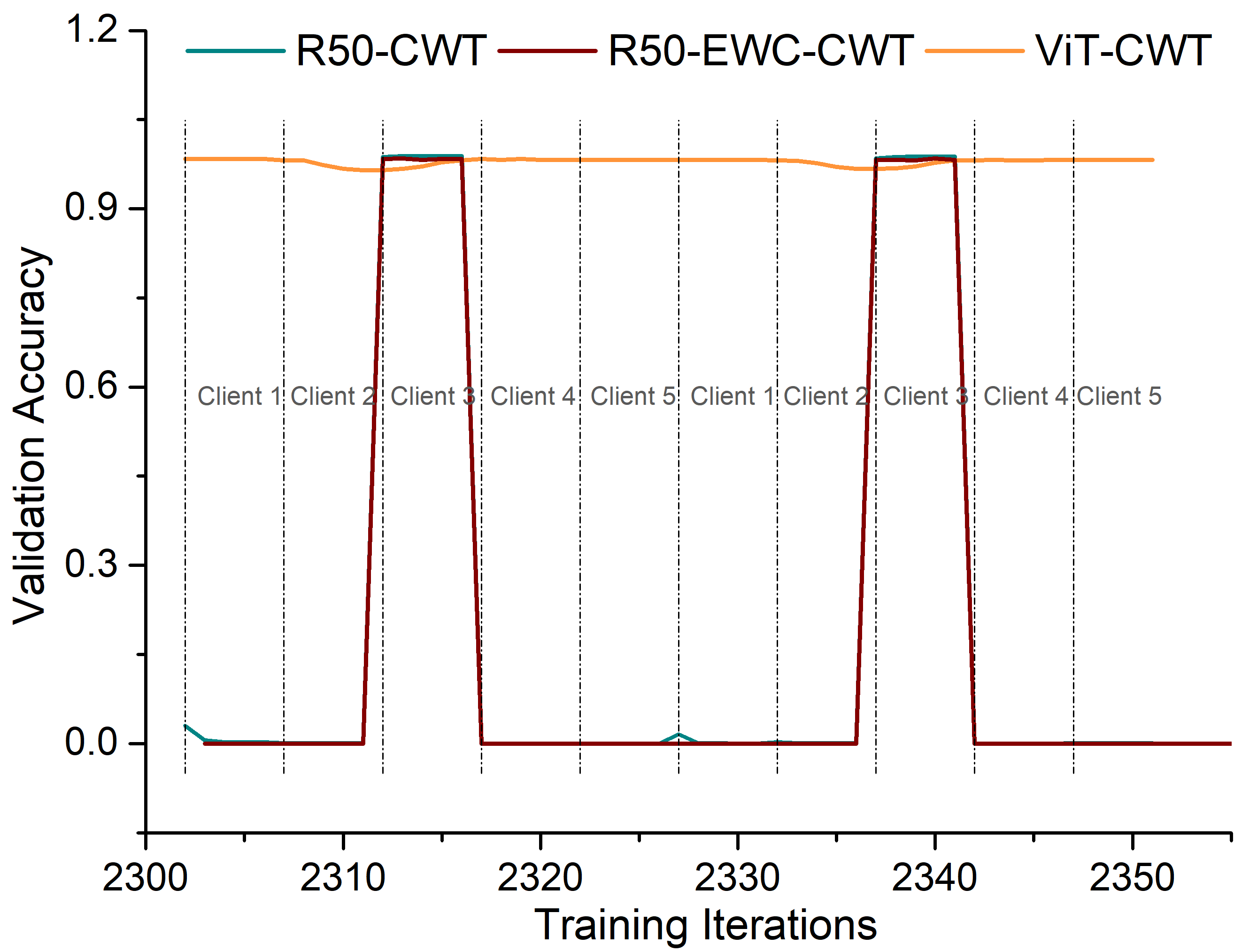}
		\end{tabular}
	\end{center}
    \vspace{-7mm}
	\caption{Left: evolution of the prediction accuracy on the validation dataset of client 3 as more clients are involved in CWT learning. We use Split 3 of CIFAR-$10$ dataset (most heterogeneous data split) and compare CWT trained with the ResNet(50) (R50 in Figure), ResNet(50)-EWC~\cite{kirkpatrick2017overcoming}, and ViT(S) models. Right: zoom in on the red rectangular in the left image. The training order of different clients is also shown. The sequential training strategy of ResNet(50)-CWT incurs catastrophic forgetting on previous clients under highly heterogeneous data distributions. ResNet(50)-EWC-CWT~\cite{kirkpatrick2017overcoming} barely solves the catastrophic forgetting problem. ViT(S)-CWT helps alleviate this problem due to its strong generalization ability and robustness to heterogeneous data.\vspace{-2mm}}
	\label{fig:catastrophe_forgetting_EWC}
\end{figure*}

\vspace{-1mm}
\subsection{Analyzing the Effectiveness of Transformers}
\label{sec:analysis}
\vspace{-1mm}

Given these promising empirical results, we now perform a careful empirical analysis to uncover what exactly contributes to Transformers' improved performance.

\vspace{-1mm}
\subsubsection{\mbox{Transformers generalize better in non-IID settings}}
\label{sec:generalize}
\vspace{-1mm}

The distributed nature of FL means that there can be substantial heterogeneity in data distributions across clients.
Prior research has shown that training FL models with FedAVG or CWT incurs issues such as weight divergence and catastrophic forgetting respectively~\cite{kirkpatrick2017overcoming,sheller2020federated}.
We argue that the local convolutions used in CNNs, which have been shown to rely more on local high-frequency patterns~\cite{geirhos2018imagenet,jo2017measuring,wang2019learning}, might be particularly sensitive to heterogeneous devices.
This problem is particularly prevalent in FL over healthcare data since input images captured by different institutions may vary significantly in local patterns (intensity, contrast, \emph{etc.}) due to different medical imaging protocols~\cite{gupta2021addressing,roth2020federated}, as well as in natural data splits due to user idiosyncrasies in speaking~\cite{latif2020federated}, typing~\cite{hard2018federated}, and writing~\cite{kairouz2019advances}.
On the other hand, ViTs use self-attention to learn global interactions~\cite{ramachandran2019stand} and have been shown to be less biased towards local patterns as compared to CNNs.
This property may contribute to their surprising robustness to distribution shifts and adversarial perturbations~\cite{bhojanapalli2021understanding,naseer2021intriguing}.
To further analyze the generalization capabilities of Transformers across heterogeneous data, we design the following experiments:

\begin{table}
\setlength\tabcolsep{4.2pt}
  \renewcommand{\arraystretch}{1}
  \centering
  \footnotesize
\caption{Prediction accuracy (\%) on a large-scale edge case setting with thousands of clients involved in training ($6,000$ and $45,000$ clients for Retina and CIFAR-$10$ respectively, with each client containing one data sample). Vision Transformers significantly outperform their ResNet counterparts in this edge case setting.\vspace{-3mm}}
  \label{table:edge_case}
  \begin{tabular}{c|cc|cc}
    \Xhline{3\arrayrulewidth}
    \centering
    &\multicolumn{2}{c|}{\textsc{Retina (\#$6,000$)}}  &\multicolumn{2}{c}{\textsc{CIFAR-10 (\#$45,000$)}} \\
    & CWT & FedAVG & CWT & FedAVG \\
    \Xhline{3\arrayrulewidth}
    ResNet(50) & $51.3 \pm 1.3$ & $55.0 \pm 0.3$ & $31.2 \pm 12.2$ & $37.5 \pm 1.4$ \\
    ViT(S) & $80.0 \pm 0.1$ & $81.0 \pm 0.1$ & $97.5 \pm 0.02$ & $97.4 \pm 0.03$ \\
    \Xhline{3\arrayrulewidth}
  \end{tabular}
\end{table}

\noindent\textbf{1. Catastrophic forgetting across heterogeneous devices:}\label{Section:catastrophic}
CNNs often work worse on out-of-distribution data. This phenomenon is especially severe in the serial FL method CWT. Due to its sequential and serial training strategy, training CNNs in a CWT paradigm usually results in catastrophic forgetting on non-IID data partitions: the model's performance on previous clients abruptly degrades after a few updates on a new client with a different data distribution~\cite{bhojanapalli2021understanding,naseer2021intriguing}. This results in poorer and slower convergence which is undesirable in FL. Similar forgetting issues have also been found in the transfer learning literature~\cite{chen2019catastrophic,chronopoulou-etal-2019-embarrassingly,serra2018overcoming}.

We evaluate CWT on Split-3 of the CIFAR-$10$ dataset to illustrate this catastrophic forgetting phenomenon. In Figure~\ref{fig:catastrophe_forgetting_EWC}, we plot the evolution of the prediction accuracy on the validation dataset of Client-3 (which shares the same data distribution as its training dataset) as more clients
are involved in CWT learning. When transferring a well-trained model on Client-3 to Client-4, the prediction accuracy on the previous Client-3 validation dataset degrades abruptly and dramatically (from $>98\%$ to $<1\%$ accuracy).
However, the model trained with ViT as backbone (ViT(S)-CWT) is able to transfer knowledge from Client-3 to Client-4 while losing only small amounts of information on Client-3 (maintains accuracy at $98\%$). Therefore, ViTs generalize better to new data distributions without forgetting old ones.

We further compare ViT(S)-CWT with an optimization method specifically designed to alleviate catastrophic forgetting, EWC~\cite{kirkpatrick2017overcoming} (using the implementation from~\cite{Hsu18_EvalCL}). Serial training of CWT on Split-3 of CIFAR-$10$ can be considered as an incremental class learning task where each client contains an exclusive subset of classes in the dataset. Each client model shares the same classifier to a standardized union label space~\cite{Hsu18_EvalCL}. However, from Figure~\ref{fig:catastrophe_forgetting_EWC}, EWC barely solves the catastrophic forgetting problem on highly heterogeneous data partitions, which also matches the results reported in~\cite{Hsu18_EvalCL}. This experiment further demonstrates the effectiveness of ViT beyond optimization methods designed for FL.

\begin{table*}
  \renewcommand{\arraystretch}{1}
  \centering
 \footnotesize
  \vspace{-3mm}
  \caption{\# transmitted message size ( \# communication round $\times$ \# model parameters (M) ) required to reach target performance (\textbf{\color{red}{best}} and \textbf{\color{blue}{second best}}).
 \# model parameters of ViT(S) and ResNet(50) is $21.7$M and $23.5$M, respectively.
  ViTs converge faster especially on heterogeneous data splits, and can be combined with optimization-based methods (FedProx and FedAVG-Share) for even faster convergence.\vspace{-2mm}}
  \label{table:communication_costs}
  \begin{tabular}{cc|cc|cccccc}
    \Xhline{3\arrayrulewidth}
    \centering
    &&\multicolumn{2}{c|}{\textsc{CWT}} &\multicolumn{6}{c}{\textsc{FedAVG}} \\
    && R50  & ViT(S) & R50 & R50-FedProx & R50-Share & ViT(S) & ViT(S)-FedProx & ViT(S)-Share \\
    \Xhline{3\arrayrulewidth}

    \multirow{3}{*}{\textsc{Retina}} & Split-1 & {\textcolor{blue}{\textbf{6 $\times$ 23.5}}}  & 9 $\times$ 21.7  & 12 $\times$ 23.5 &7 $\times$ 23.5 &  11 $\times$ 23.5 & 11 $\times$ 21.4 & {\textcolor{rr}{\textbf{4 $\times$ 21.4 }}}  & 7 $\times$ 21.4\\

    & Split-2 &72 $\times$ 23.5 & 55 $\times$ 21.4 & $\infty$ & $\infty$ & 85 & 15 $\times$ 21.4 &  {\textcolor{rr}{\textbf{12 $\times$ 21.4 }}} & {\textcolor{blue}{\textbf{13 $\times$ 21.4}}} \\

    & Split-3 &$\infty$ & 58 $\times$ 21.4 & $\infty$ & $\infty$ &  $\infty$ & {\textcolor{blue}{\textbf{15 $\times$ 21.4 }}} & {\textcolor{rr}{\textbf{12 $\times$ 21.4}}} &16 $\times$ 21.4 \\
    \Xhline{1\arrayrulewidth}

    \multirow{3}{*}{\textsc{CIFAR-$10$}} & Split-1 & 2 $\times$ 23.5 & {\textcolor{rr}{\textbf{1 $\times$ 21.4 }}} &4$\times$ 23.5  & 4 $\times$ 23.5  & 5 $\times$ 23.5& {\textcolor{rr}{\textbf{1 $\times$ 21.4 }}} & {\textcolor{rr}{\textbf{1 $\times$ 21.4 }}}  & {\textcolor{rr}{\textbf{1 $\times$ 21.4 }}}  \\

    & Split-2 & $\infty$ & 34 $\times$ 21.7 & 19 $\times$ 23.5 & 17 $\times$ 23.5 & 9 $\times$ 23.5& {\textcolor{blue}{\textbf{2 $\times$ 21.4 }}}  & {\textcolor{blue}{\textbf{2 $\times$ 21.4}}}  & {\textcolor{rr}{\textbf{1 $\times$ 21.4}}} \\

    & Split-3 & $\infty$ & 85 $\times$ 21.7 & $\infty$ & $\infty$ & 41 $\times$ 23.5 & 4  $\times$ 21.4 & {\textcolor{blue}{\textbf{3 $\times$ 21.4}}} & {\textcolor{rr}{\textbf{1  $\times$ 21.4 }}}\\
    \Xhline{3\arrayrulewidth}
  \end{tabular}
\end{table*}

\begin{figure*}
	\centering
	\begin{center}
		\begin{tabular}{cccc}
\vspace{-0.5mm}
		\qquad	\scriptsize{Split 2, KS-0.49 (Retina)} & \quad \scriptsize{Split 3, KS-0.57 (Retina)} &  \scriptsize{Split 3, KS-0.65 (CIFAR-$10$)}& \scriptsize{Split 3, KS-1 (CIFAR-$10$)} \\
	\includegraphics[width=0.245\linewidth]{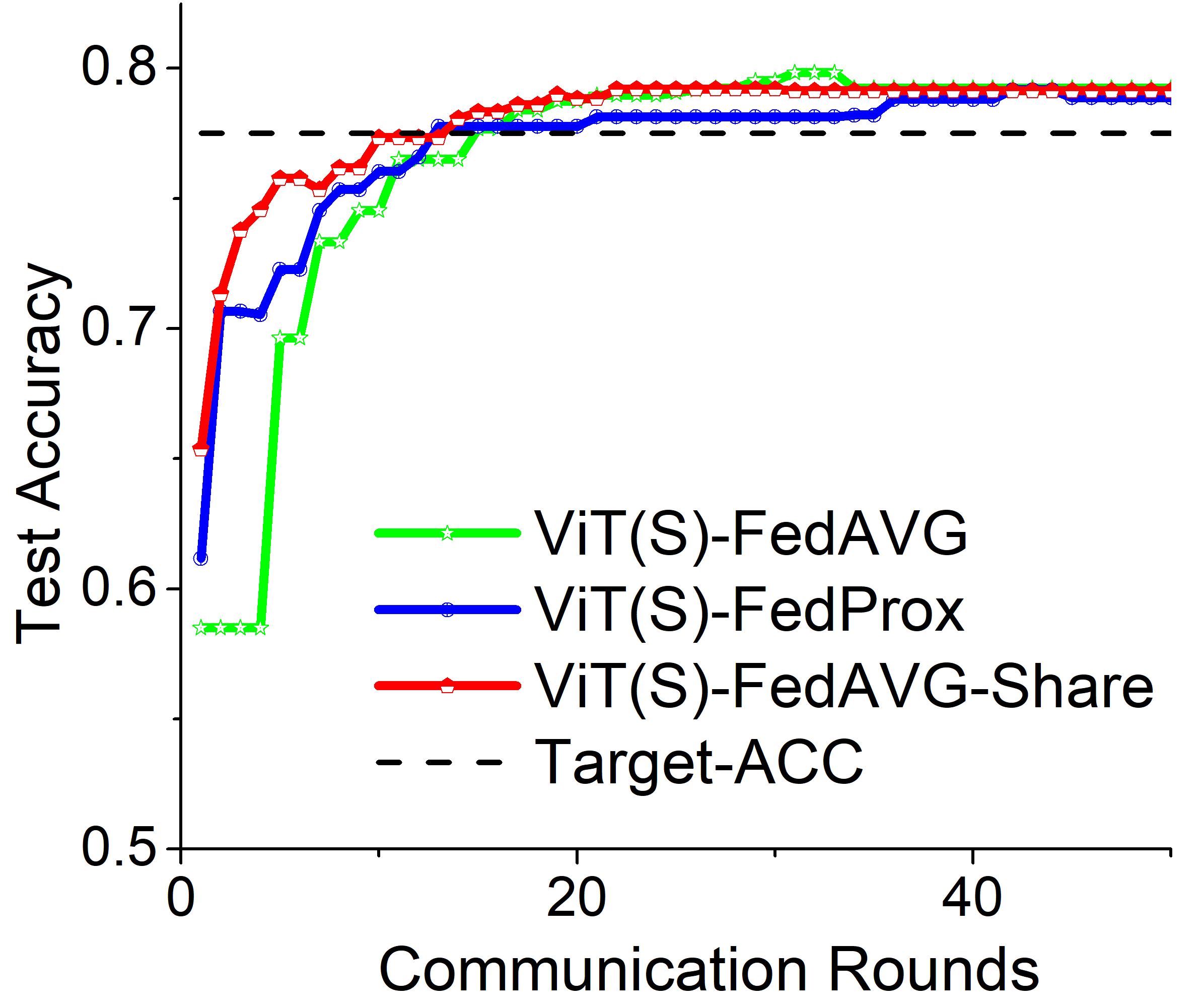}& \hspace{-3mm}
	\includegraphics[width=0.225\linewidth]{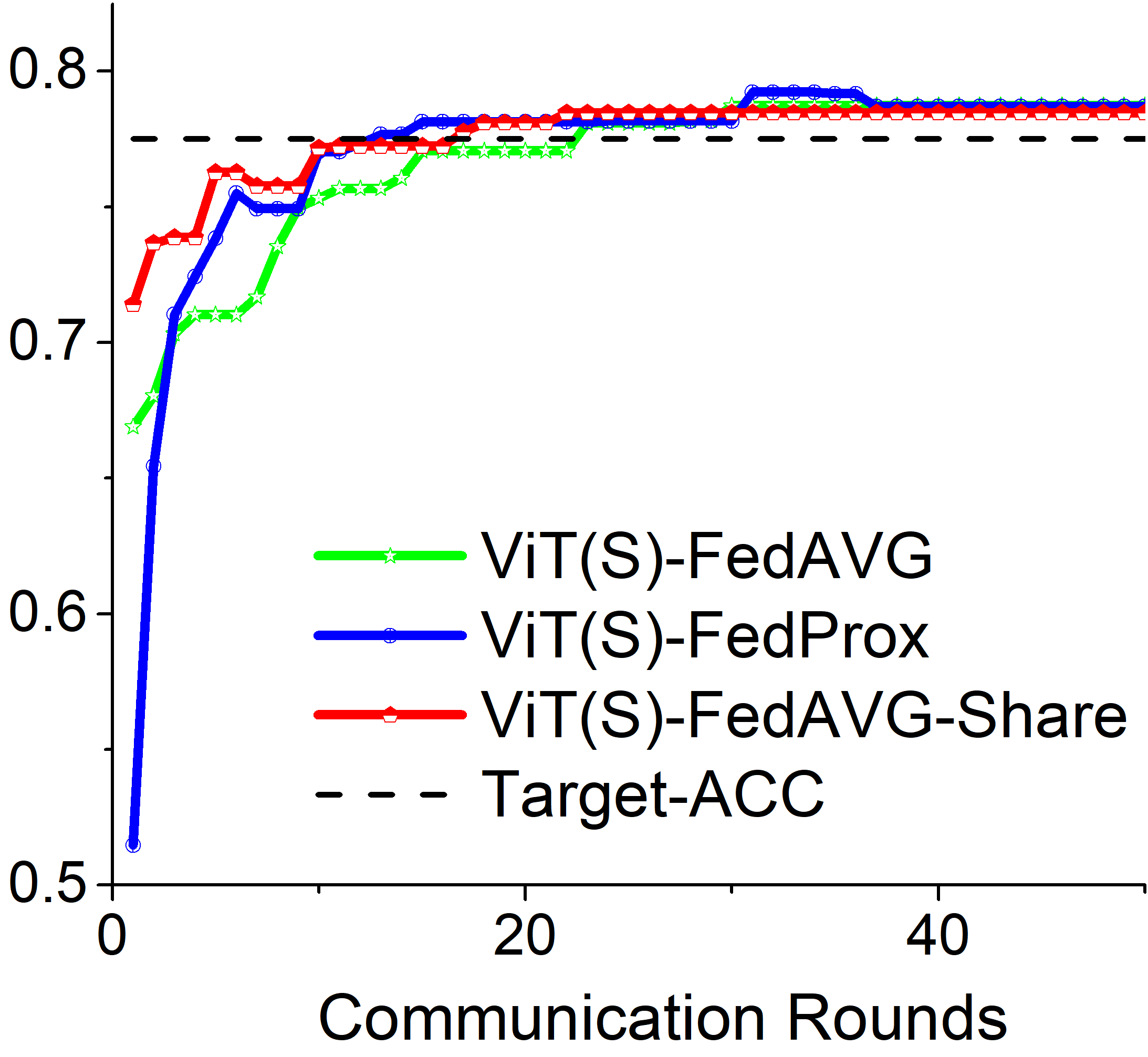}&  \hspace{-3mm}
	\includegraphics[width=0.225\linewidth]{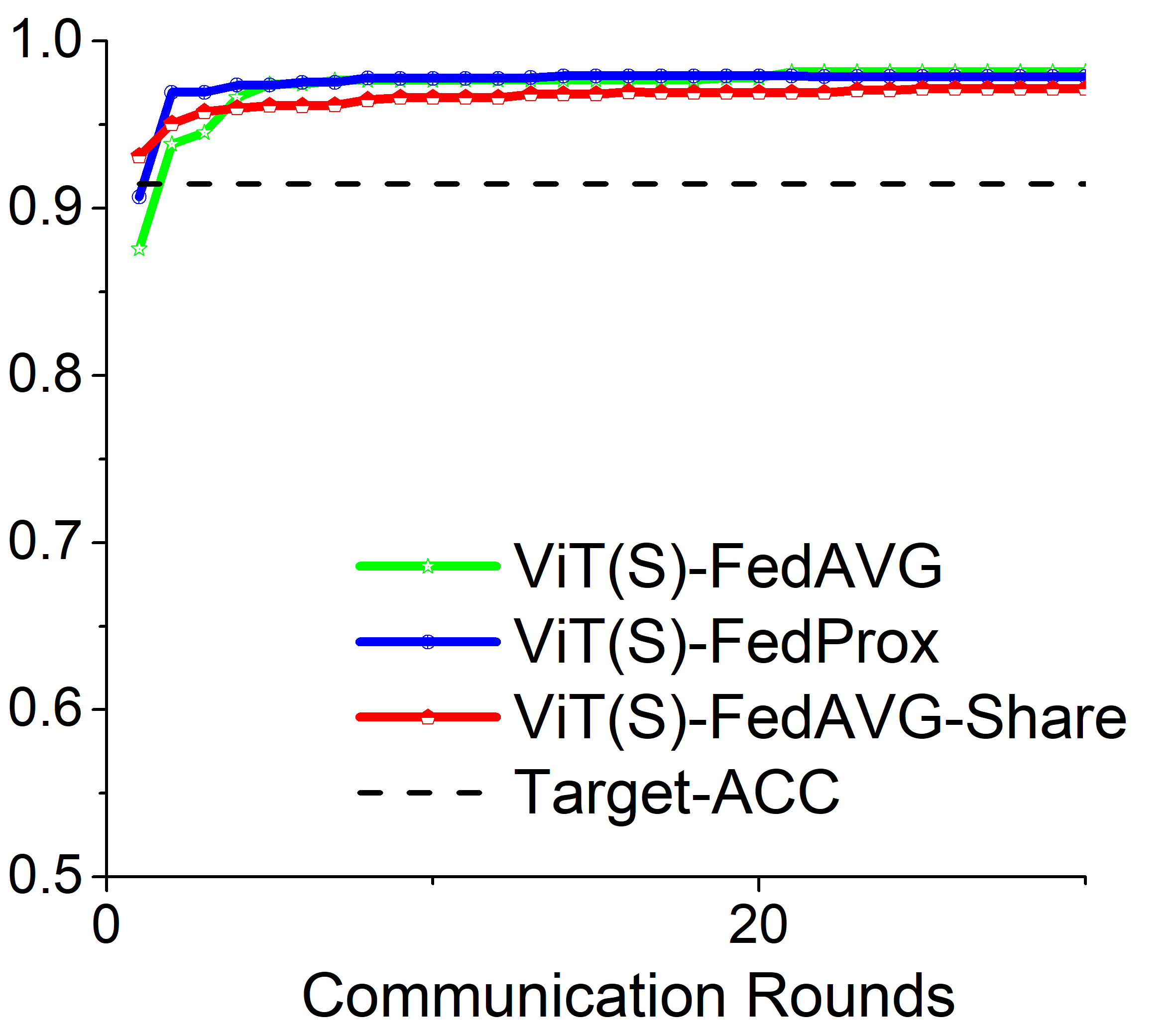}&  \hspace{-3mm}
	\includegraphics[width=0.225\linewidth]{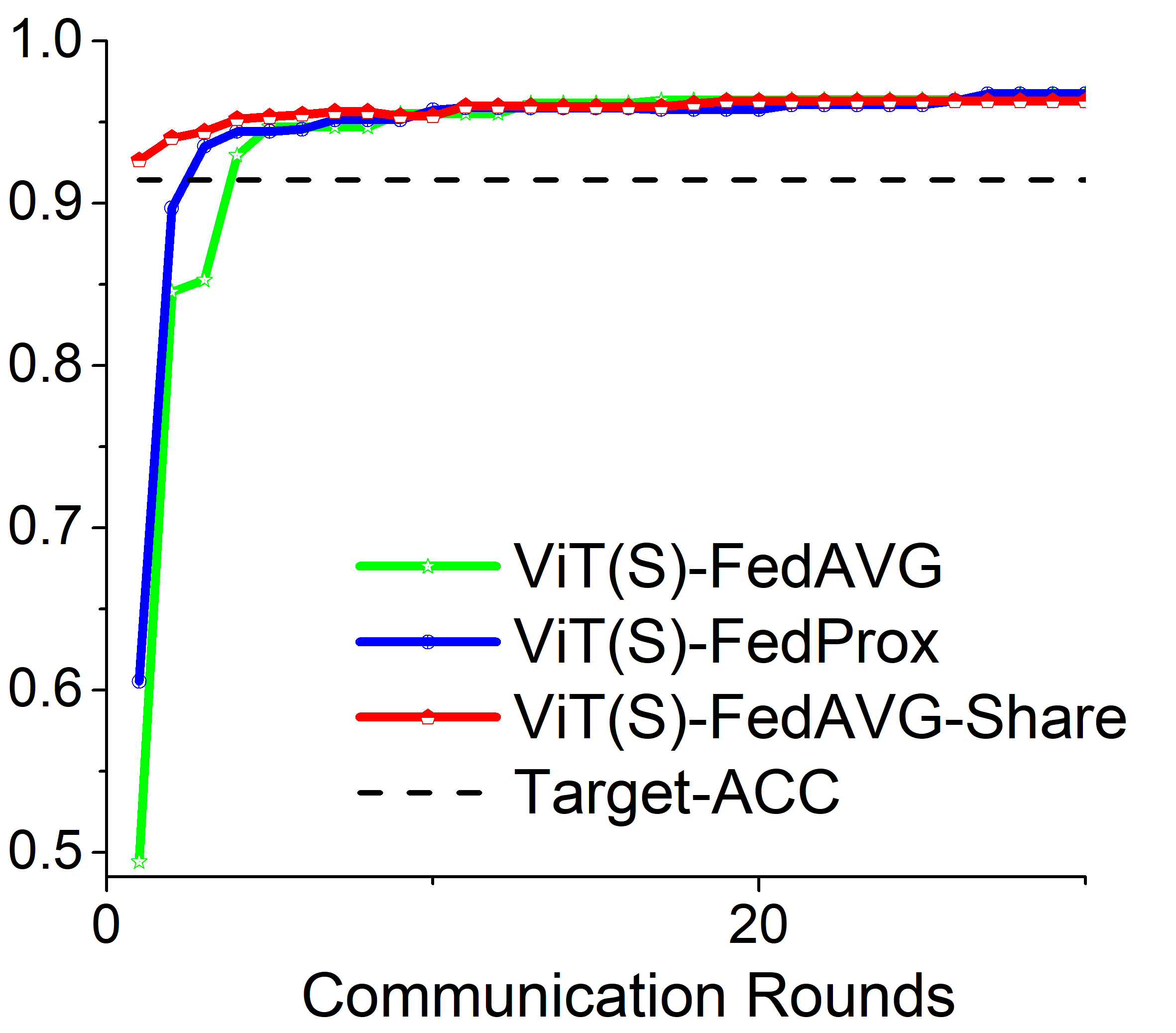}\\
		\end{tabular}
	\end{center}
    \vspace{-7mm}
	\caption{Test set accuracy versus communication rounds on ViT(S)-FedAVG and their combination with existing FL methods FedProx~\cite{li2018federated} and FedAVG-Share~\cite{zhao2018federated}. Vision Transformers can be used in conjunction with existing optimization based FL methods to further improve convergence speed and reach target performance with fewer communication rounds.}
	\label{fig:Communication_combination_FL}
\end{figure*}
\noindent\textbf{2. Generalization of \names\ on real-world federated datasets:} A well-trained federated model should perform well on out-of-distribution test datasets of other unseen clients. To test the generalizability of Transformers, we apply it to a real-world federated CelebA dataset~\cite{liu2015deep} and compare it to the ResNet counterparts, FedProx~\cite{li2018federated}, and FedAVG-Share~\cite{zhao2018federated}. We report the test accuracies of models trained using different FL methods on the union of the test data from all local clients in Table~\ref{table:celebA_results}. Our \names\ approach outperforms state-of-the-art FL methods, and also reduces variance. 
This shows that Transformers learn a better global model than their CNNs counterparts.

\noindent\textbf{3. Generalization of \names\ on extreme large-scale setting:} To validate the effectiveness of \names\ on a more large-scale real-world distributed learning setting where thousands of clients are involved, we further apply different FL methods to an extreme edge case situation on both Retina and CIFAR-$10$ dataset. The edge case here is defined as one client holding \textit{only one} data sample, which is quite common in healthcare where the patient holds only one data sample belonging to themselves. This results in an extremely large number of heterogeneous clients: $6,000$ for Retina and $45,000$ for CIFAR-$10$. From Table~\ref{table:edge_case}, ViTs still learn a promising global model on this extremely heterogeneous edge case setting, significantly outperforming ResNet models (from $50\%$ to $80\%$ on Retina and from $30\%$ to $90\%$ on CIFAR-$10$).

\vspace{-1mm}
\subsubsection{Transformers converge faster to better optimum}
\label{sec:converge}
\vspace{-1mm}

A powerful FL method should not only perform robustly on both IID and non-IID data partitions but also have low communication costs to enable deployment over communicated-limited bandwidths.
Communication cost is determined by the number of rounds till convergence and the number of model parameter. We calculate the number of communication rounds needed to achieve a predefined target test set accuracy of $95\%$ of the prediction accuracy of a centrally trained ResNet(50). Specifically, we set the target accuracy of Retina and CIFAR-$10$ dataset to be $77.5\%$ and $91.5\%$ respectively. We define one communication round on the serial CWT method as one complete training cycle across all federated local clients.

From Figure~\ref{fig:Communication_ViT_ACC} and Table~\ref{table:communication_costs}, all the evaluated FL methods converge to the target test performance quickly on homogeneous data partitions. However, the convergence speed of ResNet(50)-FedAVG and ResNet(50)-CWT decrease with increasing heterogeneity and even reach a plateau on highly heterogeneous data partitions (and never reach the target accuracy). In contrast, \names\ still converges quickly on heterogeneous data. For example, ResNet(50)-CWT completely diverges due to severe catastrophic forgetting on heterogeneous data partitions Split-2 and Split-3 on CIFAR-$10$, whereas ViT(S)-CWT reaches the target performance after $34$ and $85$ communication rounds.



\begin{figure*}
	\centering
	\begin{center}
		\begin{tabular}{ccc}
		\qquad	\scriptsize{Split 1, KS-0 (Retina)} &  \scriptsize{Split 3, KS-0.57 (Retina)} \vspace{-1mm} &  \scriptsize{Split 3, KS-1 (CIFAR-$10$)} \\
        \includegraphics[width=0.29\linewidth]{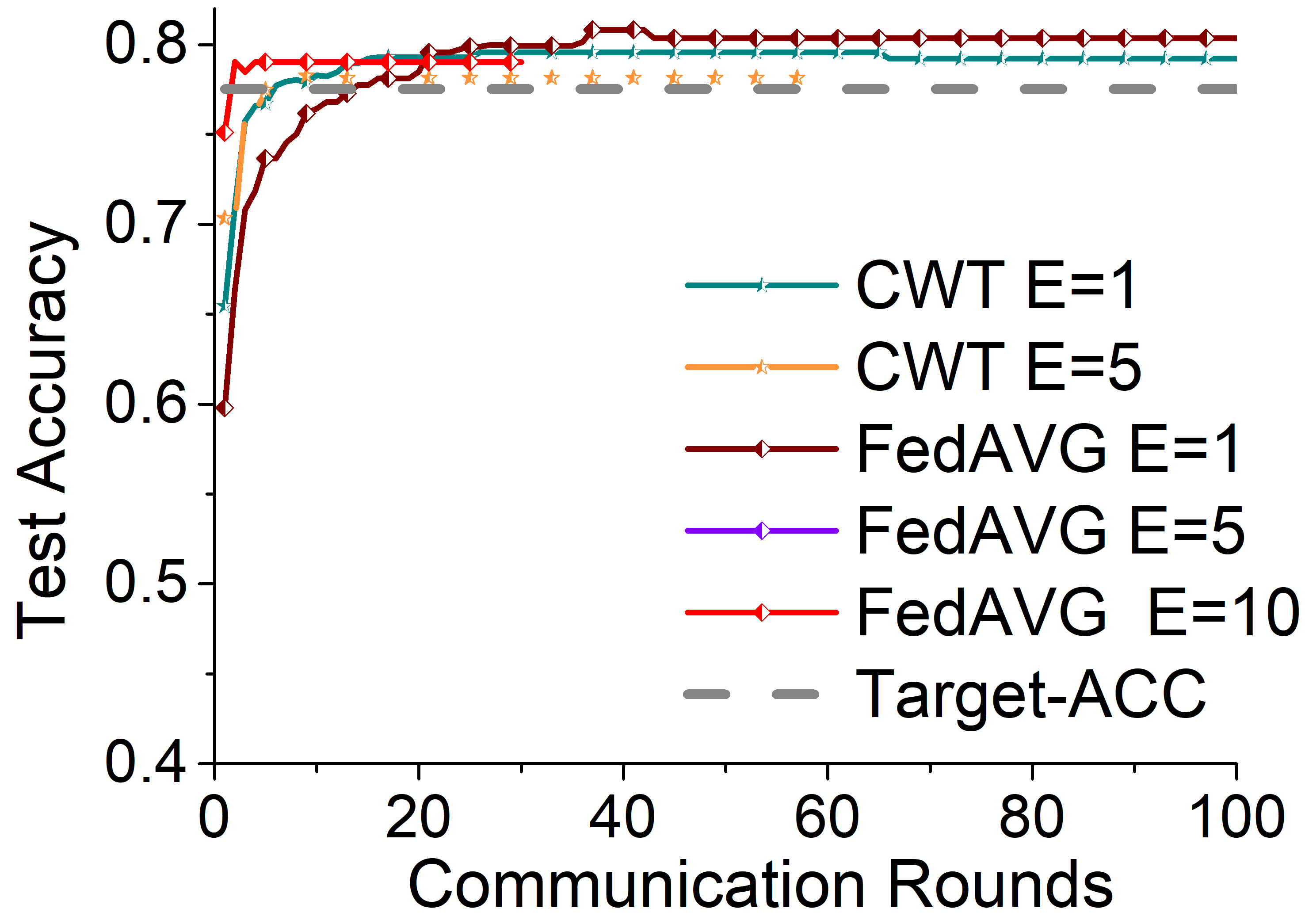} &
        \includegraphics[width=0.27\linewidth]{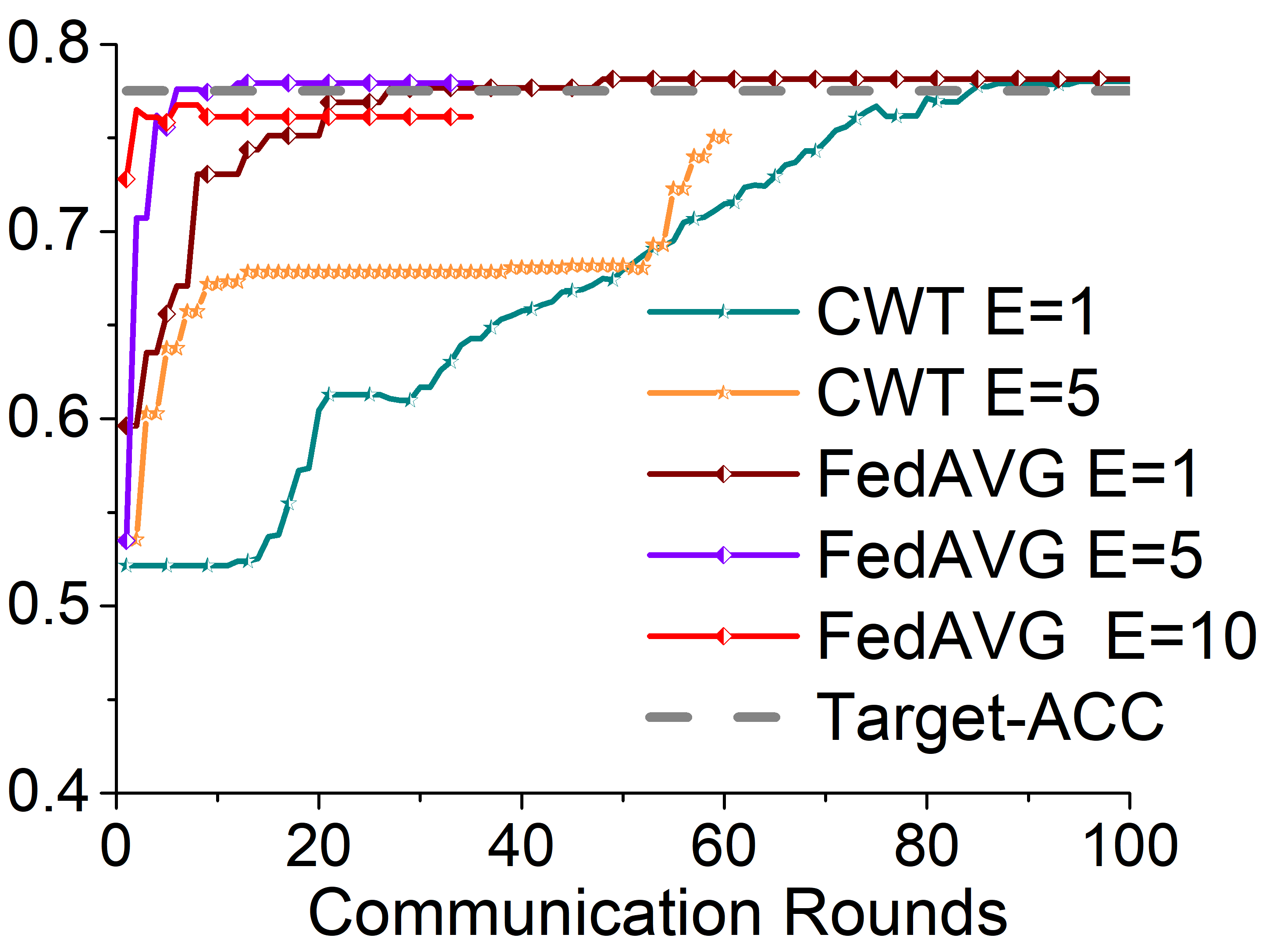} &
        \includegraphics[width=0.27\linewidth]{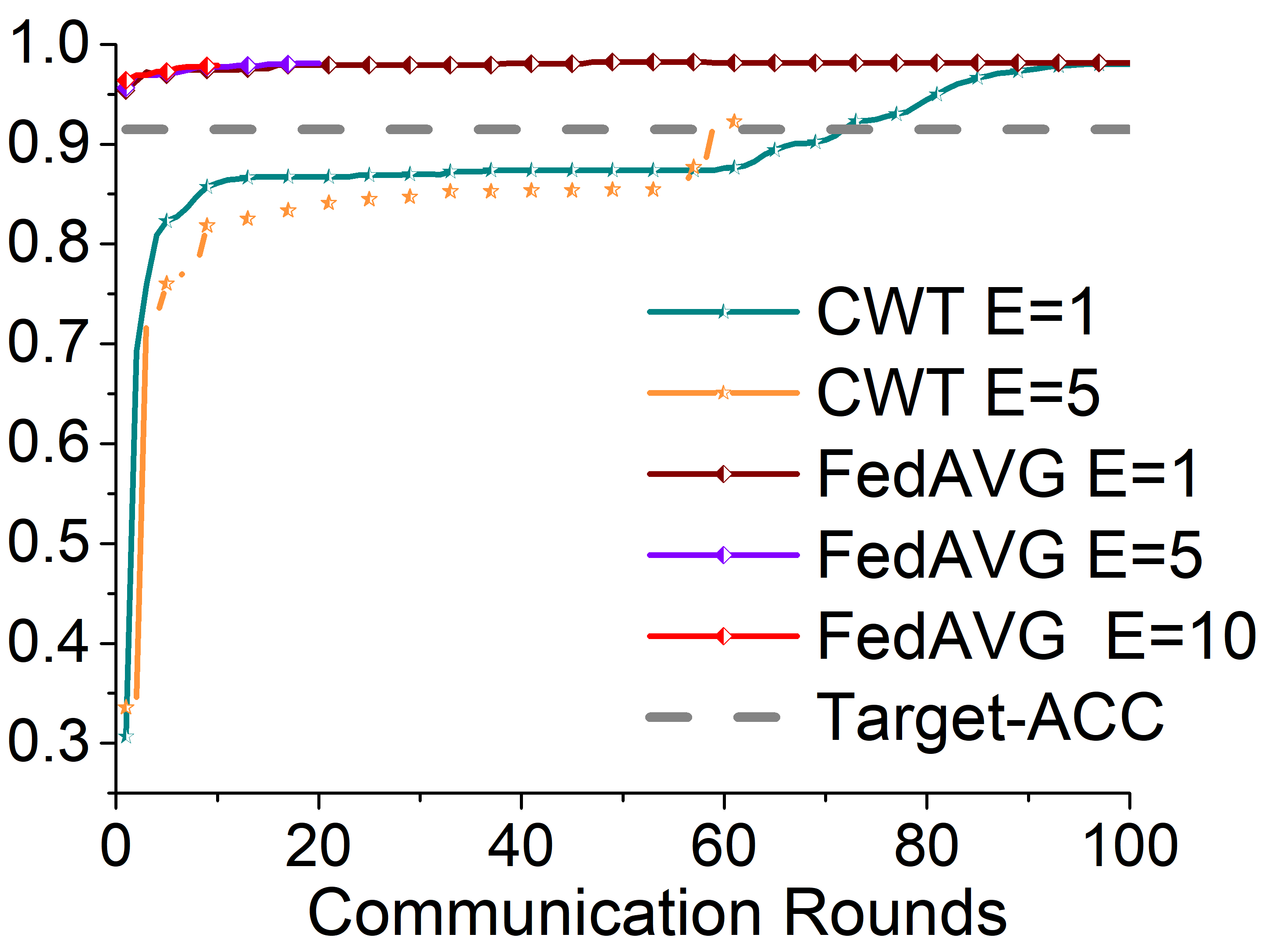} \\
        \end{tabular}
	\end{center}
	\vspace{-6mm}
	\caption{The effect of training over different local epochs $E$ on each communication round for ViT(B)-CWT and ViT(B)-FedAVG models on Retina and CIFAR-$10$ (the ViT(B) prefix of CWT and FedAVG in the legend labels is omitted for simplicity). Large $E$ leads to faster convergence in mild heterogeneous data partitions, but might lead to worse final performance in severely heterogeneous data partitions.}
    \label{fig:ViT_ACC_Epochs}
\end{figure*}

\vspace{-1mm}
\subsection{In Conjunction with Existing Methods}
\label{sec:applied_to_FLs}
\vspace{-1mm}

Since our investigation into architectural choices is largely orthogonal to existing optimization based FL methods, our findings can be easily used in conjunction with the latter. We combine Vision Transformers with optimization-based methods (FedProx~\cite{li2018federated} and FedAVG-Share~\cite{zhao2018federated}), and apply it to both Retina and CIFAR-$10$ datasets. From Table~\ref{table:communication_costs} and Figure~\ref{fig:Communication_combination_FL}, when applying to existing FL optimization methods, \ViT\ further boosts the performance for heterogeneous data clients. 

\vspace{-1mm}
\subsection{Take-aways for Practical Usage}
\label{sec:tips}
\vspace{-1mm}

\textbf{Local training epochs:} It is standard to use $E$ to denote the number of rounds a local model passes over its local dataset. $E$ is known to strongly affect the performance of FedAVG \cite{mcmahan2016communication} and CWT \cite{chang2018distributed}. We conduct an experimental study on the impact of local training epochs $E$ on \names. We consider $E\in \{1, 5, 10\}$ for ViT(B)-FedAVG, and $E\in \{1, 5\}$ for ViT(B)-CWT. From Figure~\ref{fig:ViT_ACC_Epochs}, we find that ViT shows similar phenomena to their CNN counterparts, \emph{i.e.}, larger $E$ accelerates convergence of ViT(B)-FedAVG on homogeneous data partitions, but may lead to deterioration of final performance on heterogeneous data partitions.
Similarly, ViT(B)-CWT also favors frequent transfer rate between each client as ResNet(50)-CWT~\cite{chang2018distributed} on non-IID data partitions. Therefore, we suggest users apply large $E$ on homogeneous data to reduce communication, but a small $E$ ($E \leq$ 5 for \ViT-FedAVG and $E=1$ for \ViT-CWT) for highly heterogeneous cases.

\begin{table}
  \renewcommand{\arraystretch}{1}
  \centering
  \footnotesize
  \caption{The influence of pretraining of Swin(T)-FedAVG on CIFAR-$10$. Similar to the training of \ViT, pretraining is important for the training of \names.\vspace{-3mm}}
  \label{table:pretrain_influence}
  \begin{tabular}{c|cccc}
    \Xhline{3\arrayrulewidth}
    \centering
    & \textsc{Central} & \textsc{Split-1} & \textsc{Split-2} & \textsc{Split-3} \\
  \Xhline{3\arrayrulewidth}
  Pretrain & $97.91$ & $98.17$ & $97.78$ & $96.40$ \\
  From scratch  & $94.50$ & $86.91$ & $79.43$ & $64.50$ \\
  \Xhline{3\arrayrulewidth}
  \end{tabular}
  \vspace{-2mm}
\end{table}

\textbf{The influence of pretraining on \textsc{ViT-FL}:}
Evidence suggests that \ViT\ generally require a larger amount of training data to perform better than CNNs when trained from scratch~\cite{dosovitskiy2020image}. We conduct experiments to investigate the influence of pretraining on \names. We apply FedAVG as the training algorithm, use Swin(T)~\cite{liu2021swin} as the backbone network, and test on CIFAR-$10$. We apply the same augmentation and regularization strategies as~\cite{liu2021swin} during training and set the maximum communication rounds to $300$. As shown in Table~\ref{table:pretrain_influence}, the performance of Swin(T) drops when trained from scratch for both the ideal centrally-hosted and FL settings. Despite this, its performance on highly-heterogeneous data partition Split-3 when trained from scratch ($64.50\%$) is surprisingly better than ResNet(50)-FedAVG ($59.68\%$ on Figure~\ref{fig:fig_acc_on_Cifar10_Retina}) when pretrained with orders of magnitude more data. In real applications, users are recommended to apply \ViT\ as their first option, since \names\ consistently outperform their CNNs counterparts when pretrained models are applied (Figure~\ref{fig:acc_vs_parameters} and Figure~\ref{fig:fig_acc_on_Cifar10_Retina}). If large-scale pretraining datasets are not available, self-supervised pretraining~\cite{caron2021emerging,he2021masked} could be an alternative.

\textbf{Other training tips:} The training strategy of \ViT\ in FL can be directly inherited from \ViT\ training, such as using linear warm-up and learning rate decay, and gradient clipping. Relatively small learning rates and gradient norm clip are necessary to stabilize the training of \ViT\ in CWT, especially in highly heterogeneous data partitions. Gradient norm clip also helps in the training of FL with CNNs across heterogeneous data since it has been shown to reduce weight divergence between local updates and the current global model~\cite{li2018federated}. Please refer to Appendix~\ref{appendix:tips} for more general tips and experimental analysis.

\vspace{-1mm}
\section{Conclusion}
\vspace{-1mm}

Despite the recent progress in FL, there remain challenges regarding convergence and forgetting when dealing with heterogeneous data.
Unlike previous methods on improving optimization, we provide a new perspective by rethinking architecture design in FL.
Using the robustness of Transformers to heterogeneous data and distribution shifts, we perform extensive analysis and demonstrate the advantages of Transformers in alleviating catastrophic forgetting, accelerating convergence, and reaching a better optimum for both parallel and serial FL methods. We release our code and models to encourage developments in robust architectures in parallel to efforts on the optimization front.

\vspace{-1mm}
\section*{Acknowledgments}
\vspace{-1mm}
This work was supported in part by a grant from the NCI, U01CA242879.

\clearpage

\clearpage
\appendix

\section*{Appendix}

\vspace{-1mm}
\section{Experimental Details}
\label{appendix:details}
\vspace{-1mm}
\begin{figure*}[t]
	\centering
	\begin{center}
		\begin{tabular}{cc}
\vspace{1mm}
      {\hspace{-10mm} (a) Split 2, KS-0.49 } & \hspace{-10mm}   {(b) Split 3, KS-0.57} \\
 \vspace{-2mm}
        \includegraphics[width=0.4\linewidth]{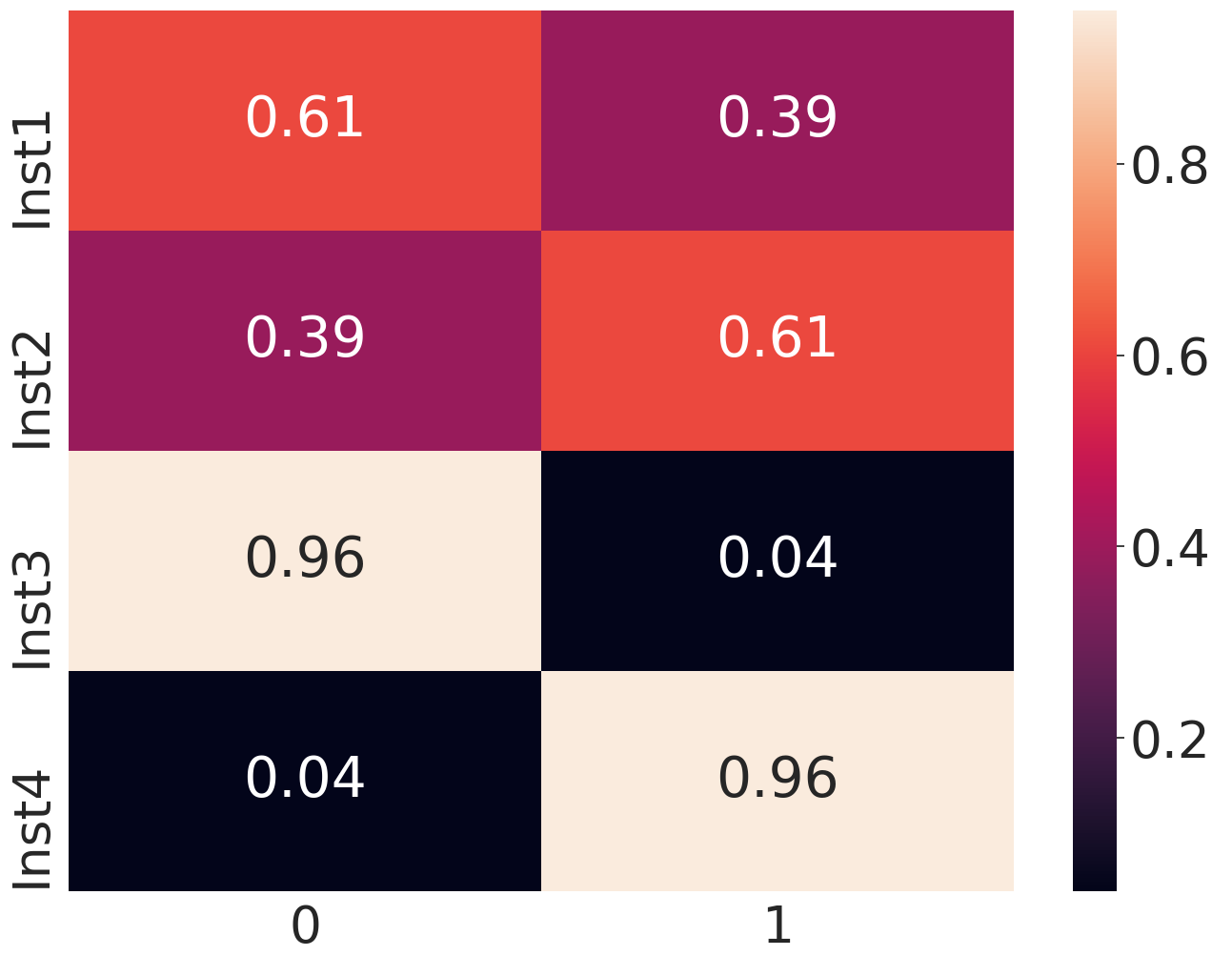} &
        \includegraphics[width=0.4\linewidth]{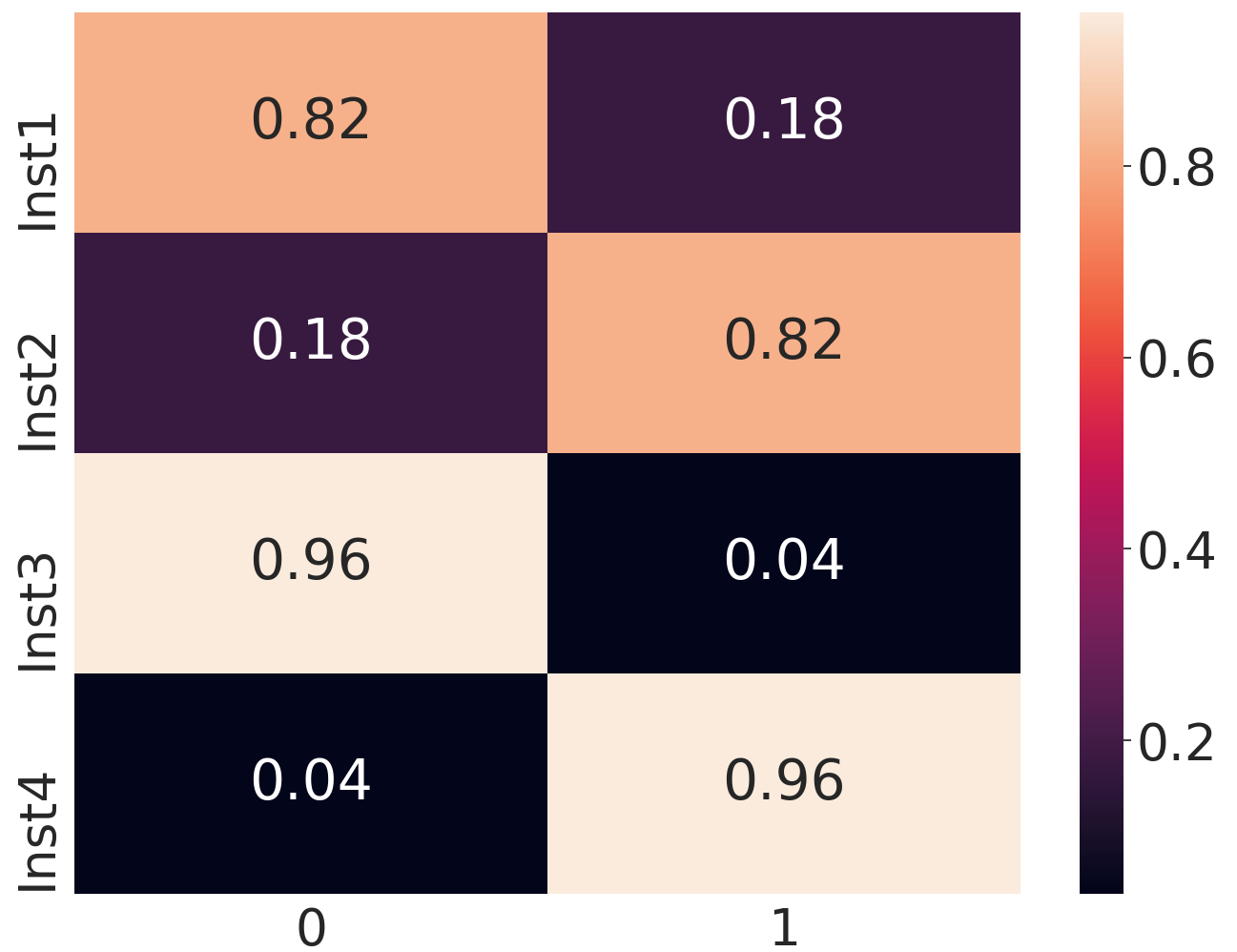}\\
\hspace{-10mm}     {Labels}  &  \hspace{-10mm}  {Labels} \\

        \end{tabular}
	\end{center}
	\caption{Detailed non-IID data partitions on \textsc{Retina} with label distribution skew. The value in each rectangle shows the fraction of data samples of a class over their total number.}
    \vspace{-3mm}
    \label{fig:data_partitions_retina}
\end{figure*}

In this section, we provide additional details on the datasets used, preprocessing steps, and experimental methodology. We include code to reproduce our experiments at \url{https://github.com/Liangqiong/ViT-FL-main}.

\subsection{Detailed Image Pre-processing and Data Partitions}
\label{appendix:partition}

\textbf{Kaggle Diabetic Retinopathy competition} (\textsc{Retina})~\cite{KaggleRetina} contains a total of $17,563$ pairs of right and left color digital retinal fundus images. Each image is labeled on a scale of $0$ to $4$ by a well-trained clinician, indicating no, mild, moderate, severe, and proliferative diabetic retinopathy respectively. Following~\cite{chang2018distributed}, we exclude the samples with scale $1$, and then binarize the remaining labels to Healthy (scale $0$) and Diseased (scale $2, 3$ or $4$). Furthermore, we only use the left images in our study to remove the confounding factor of different disease status of left and right eyes for the same patient. We randomly select $6,000$ balanced ($3,000$ healthy and $3,000$ diseased) images for training, $3,000$ balanced images as the global validation dataset, and $3,000$ balanced images as the global test dataset. Other image pre-processing steps include rescaling as a radius of $300$, local color averaging and image clipping, resizing to $256 \times 256$, horizontal flipping, and randomly cropping to $224 \times 224$. We choose a final $224 \times 224$ image dimension to be compatible with current work in both CNNs \cite{he2016deep} and Vision Transformers \cite{dosovitskiy2020image}.

\begin{figure*}[t]
	\centering
	\begin{center}
		\begin{tabular}{cc}
\vspace{1mm}
      {\hspace{-10mm} (a) Split 2, KS-0.65 } & \hspace{-10mm}   {(b) Split 3, KS-1} \\
        \includegraphics[width=0.42\linewidth]{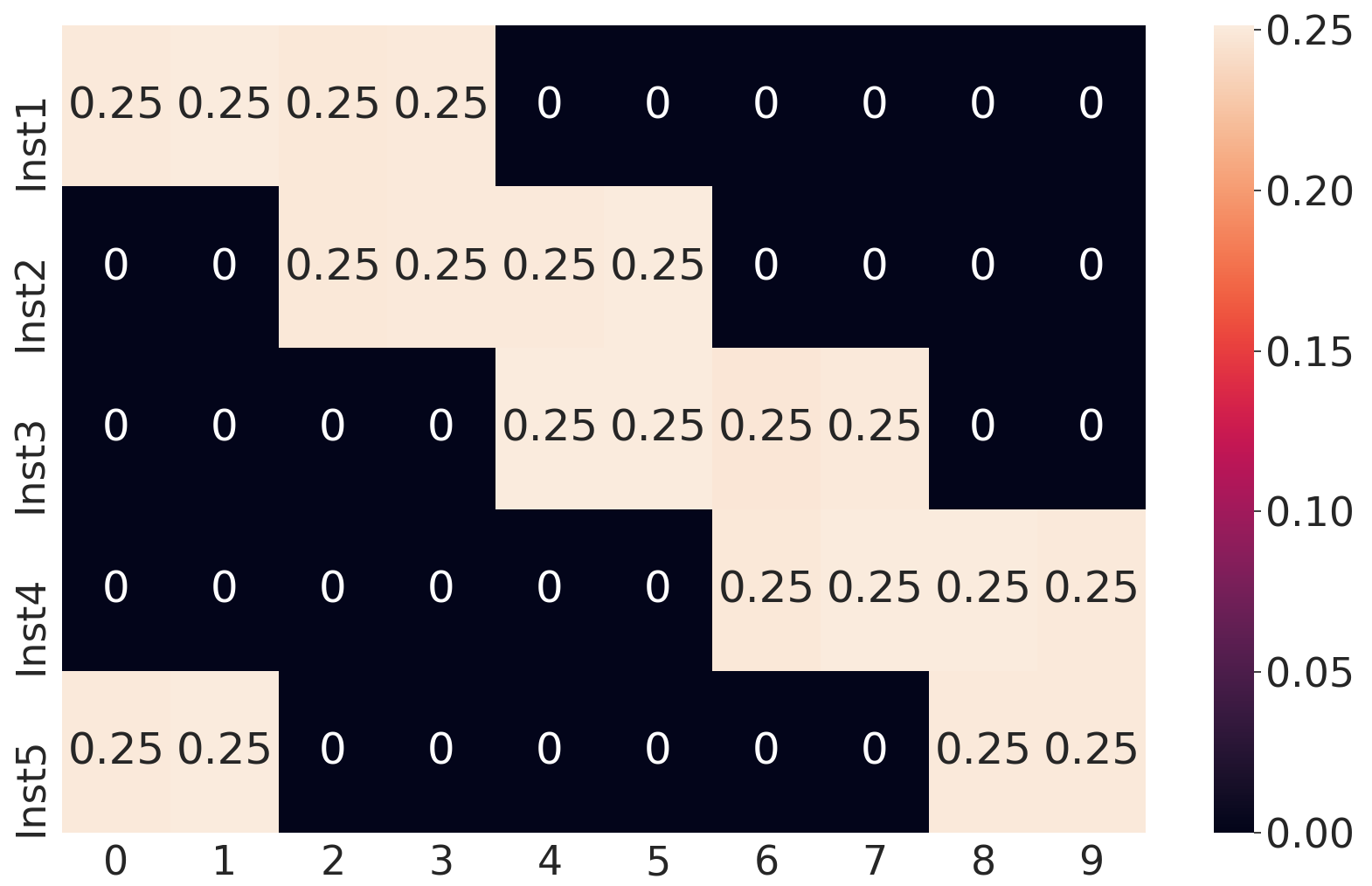} &
        \includegraphics[width=0.42\linewidth]{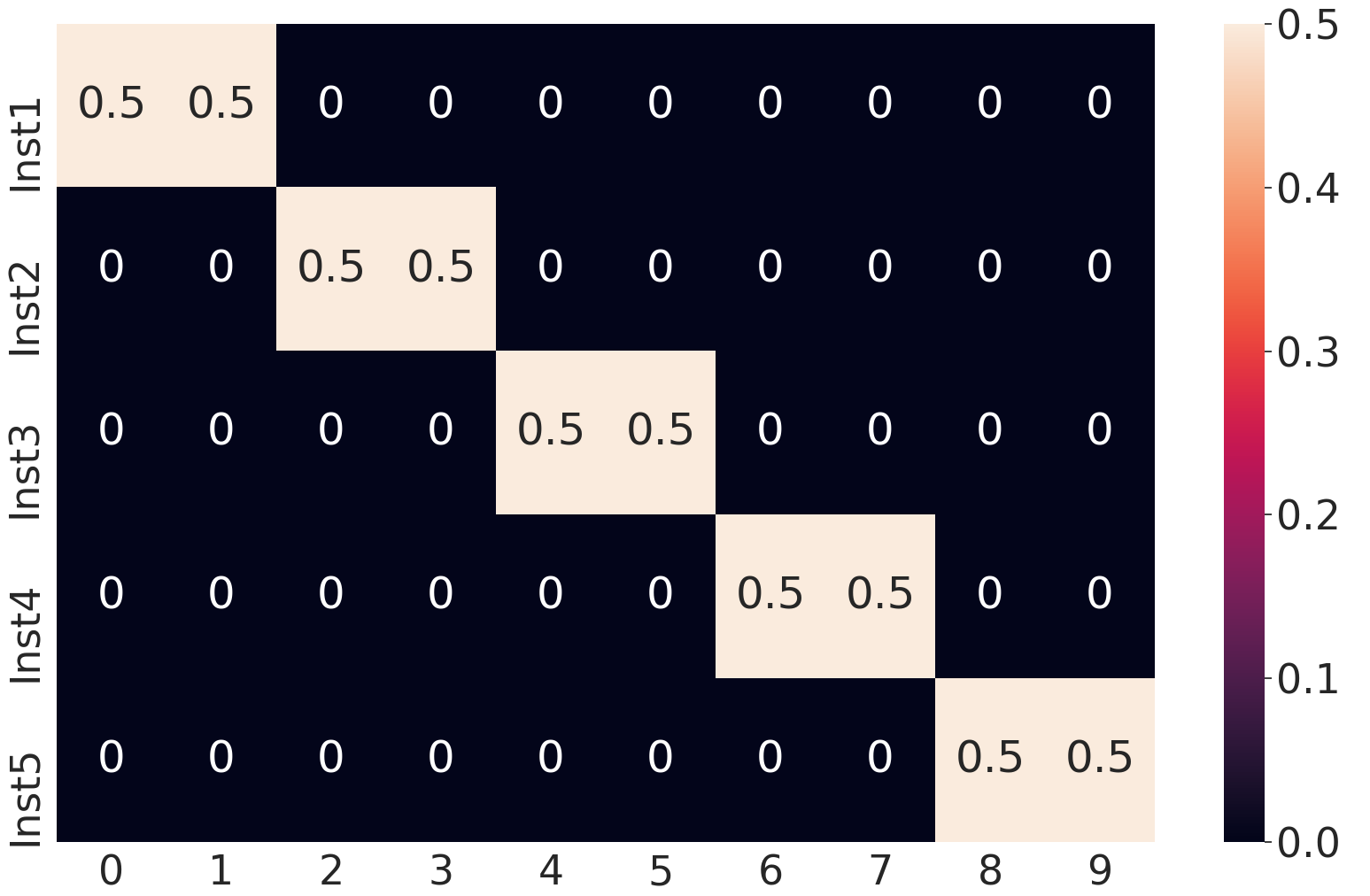}\\
\hspace{-10mm}     {Labels}  &  \hspace{-10mm}  {Labels} \\

        \end{tabular}
	\end{center}
	\caption{Detailed non-IID data partitions on \textsc{CIFAR-$10$} with label distribution skew. The value in each rectangle shows the fraction of data samples in a class over their total number.}
    \vspace{-3mm}
    \label{fig:data_partitions_cifar}
\end{figure*}

\begin{figure*}[t]
	\centering
	\begin{center}
		\begin{tabular}{cc}
			\qquad \scriptsize{Split 3, KS-0.57 (Retina)} & \hspace{-28mm} \scriptsize{Split 3, KS-1 (CIFAR-$10$)} \\
        \includegraphics[width=0.4\linewidth]{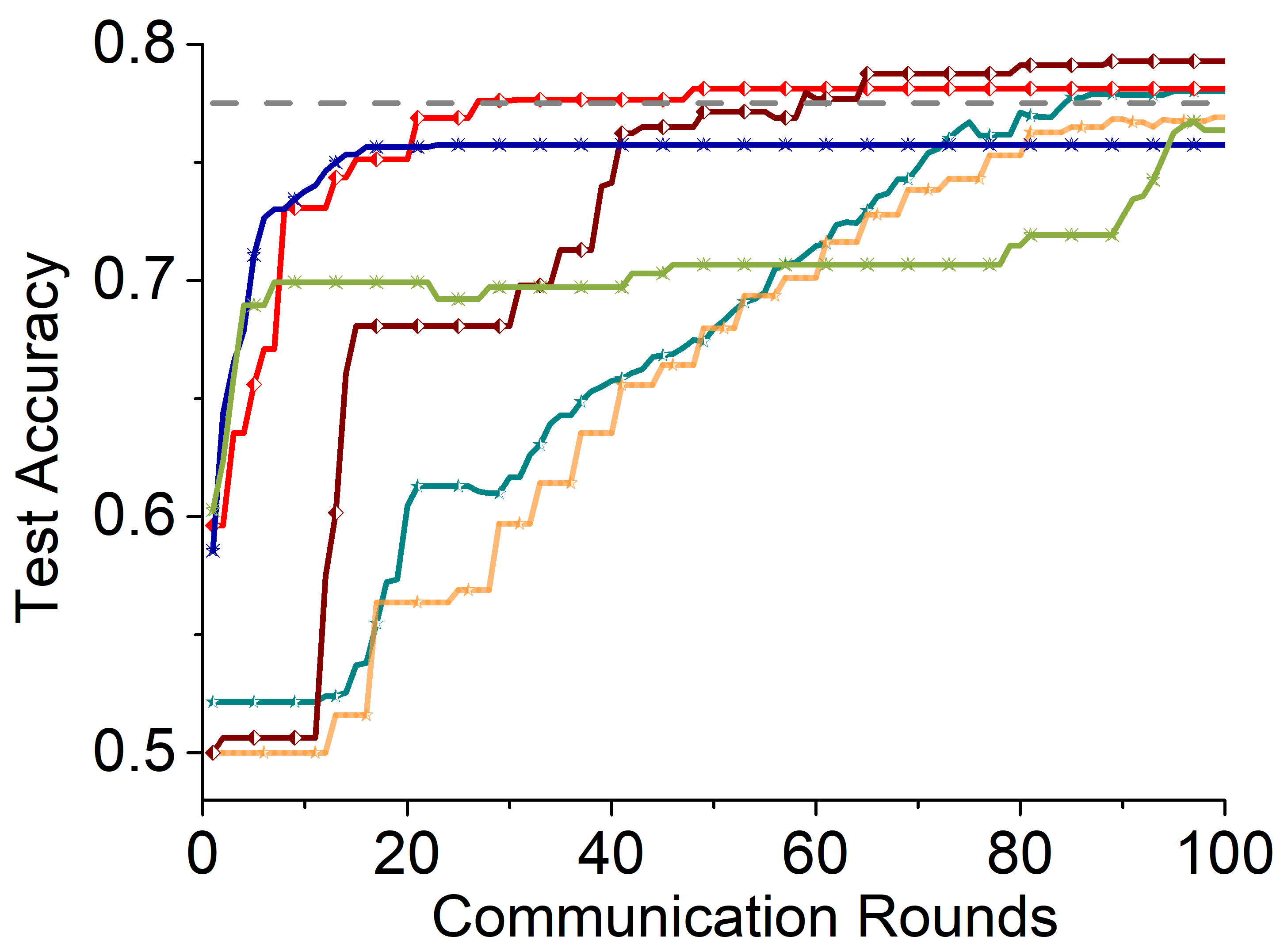} &
        \includegraphics[width=0.59\linewidth]{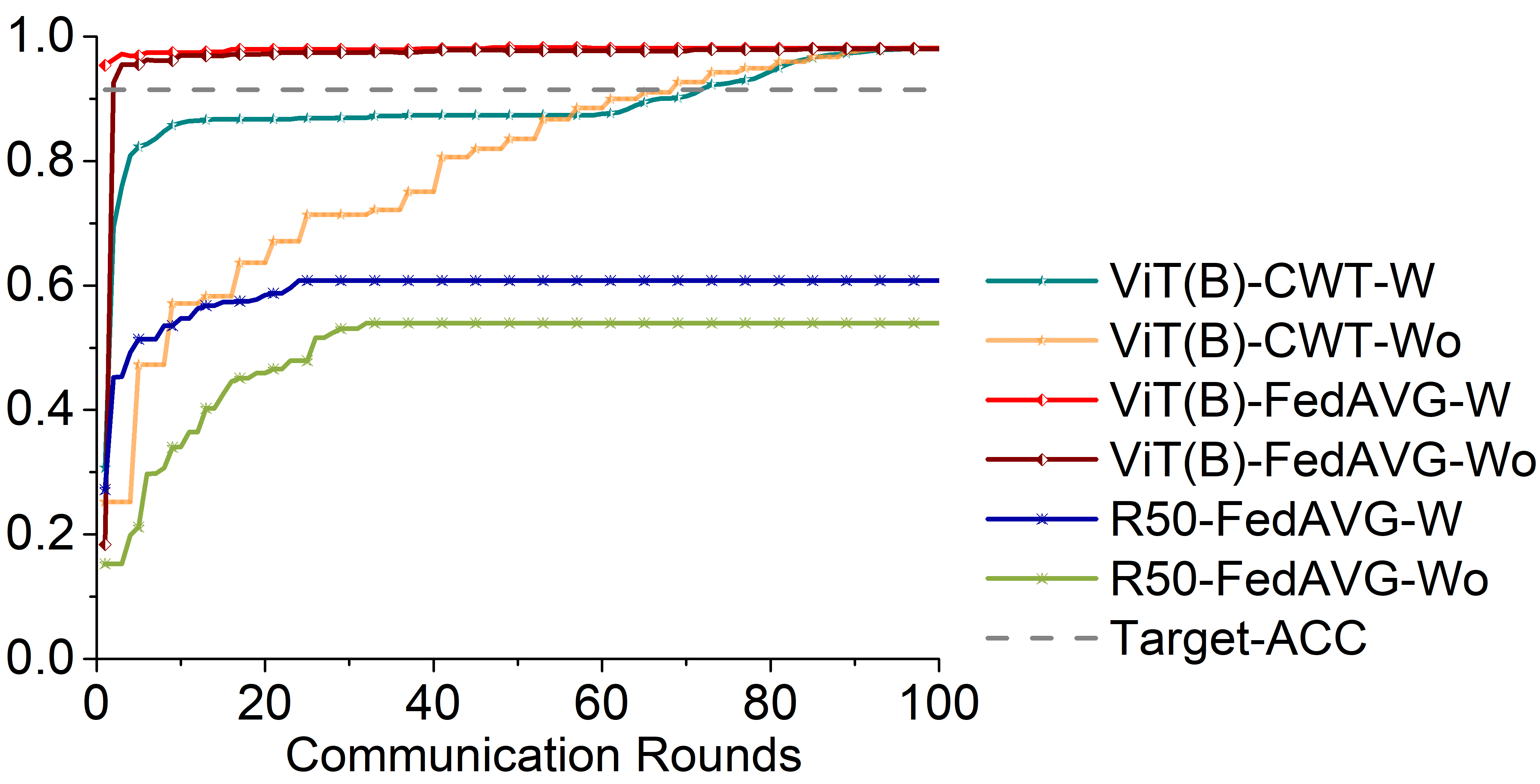} \\
        \end{tabular}
	\end{center}
	\vspace{-3mm}
	\caption{Influence of gradient clip on different FL methods with ViT(B) and ResNet-50(R50) as the backbone networks, respectively.  In the legend, \textbf{-W} denotes with gradient clip and \textbf{-Wo} denotes without gradient clip. We find that gradient clip stabilizes training and accelerates convergence speed on highly heterogeneous data splits.}
    \vspace{-2mm}
    \label{fig:gradient_clip}
\end{figure*}

We simulate three sets of data partitions for the \textsc{Retina} dataset with each data partition containing four simulated clients: one IID-data partition (Split 1, KS-0), and two non-IID data partitions with label distribution skew (Split 2, KS-0.49, and Split 3, KS-0.57). See Figure~\ref{fig:data_partitions_retina} for the detailed non-IID data partitions. 

\textbf{CIFAR-10}~\cite{krizhevsky2009learning} consists of $50,000$ training and $10,000$ testing $32 \times 32 $ images in $10$ classes, with $5,000$ and $1,000$ images per class in training and test dataset respectively. Following~\cite{hsieh2020non}, we apply the $10,000$ image test dataset as the global test dataset, set aside $5,000$ images from the training dataset as the global validation dataset, and the remaining $45,000$ images as training dataset. We preprocess each image by resizing to $256 \times 256$ and cropping to $224 \times 224$.

We simulate one IID-data partition (Split 1, KS-0), one heterogeneous data partition (Split 2, KS-0.65), and one heterogeneous data partition in the extreme case (Split 3, KS-1). Each data partitions contains five clients~\cite{chang2018distributed}. We randomly assign each client with images sampled via a uniform distribution over the $10$ classes for the IID data partition Split 1, KS-0. For Split 2, KS-0.65, one client receives images sampled from two classes, while the remaining four clients receive images sampled from four classes. Split 3, KS-1 is an extreme case where each client receives images sampled from only two classes. Please refer to Figure~\ref{fig:data_partitions_cifar} for the detailed label distribution on each client for Split 2 and Split 3. 

\textbf{CelebA} is a large-scale face attributes dataset with more than $200$K celebrity images. The images in CelebA cover large diversities, \emph{i.e.}, large pose variations and background clutter. We use a specially designed federated version of CelebA provided by the LEAF benchmark~\cite{caldas2018leaf} which partitions the dataset into devices based on the celebrity in the picture (\emph{i.e.}, each device contains only images of celebrity). Following~\cite{caldas2018leaf}, we test on the binary classification task (presence of smile),  drop clients with larger than 8 samples to increase the difficult. This results in a total of $227$ clients with an average of $5.34\pm1.11$ samples and a total of $1213$ samples. for the histogram of the number of training samples in each client. We preprocess each image by resizing to $256 \times 256$ and cropping to $224 \times 224$.

\subsection{Implementation Details and Hyperparameters}
\label{appendix:hyperparameter}





\textbf{Implementation Details.} All the methods are implemented with Pytorch and optimized either with SGD (with momentum as $0.9$ and no weight decay) or AdamW~\cite{kingma2014adam} (with weight decay as 0.05). All experiments were conducted on either a TITAN V GPU or GeForce RTX 2080 GPU. For fair comparison, all the models used in this paper are pretrained from ImageNet’s ILSVRC-2012~\cite{deng2009imagenet}. We set local training epoch in all the FL methods to $1$, and the total communication round to 100, unless otherwise stated. We set the local training batch size to $32$, and adopt a default input image resolution $224 \times 224$ for all methods. More implementation details are shown below.

\textbf{Training hyperparameters:} Inherited from original Transformers training, the Swin-FL models are optimized with AdamW~\cite{kingma2014adam}, and the ViT-FL models are optimized with SGD. As a fair comparison, the optimizers for the compared CNNs are selected from either SGD and AdamW according to parameter searching. We use linear learning rate warm-up and decay scheduler for the Transformer models. Specifically, we set the warmup steps to $500$, and cosine learning rate decay to zero after the maximum round of FL training epochs is reached. The learning rate scheduler for FL with CNNs is selected from linear warm-up and decay or step decay (halved every $30$ rounds of FL training). Gradient clipping (at global norm 1) is applied to stabilize the training.

\textbf{Hyperparameter selection:} We tune the best parameters (including learning rate scheduler, and penalty constant $\mu$ in the proximal term of FedProx) for FL with CNNs on Split-2 of \textsc{Retina} and \textsc{CIFAR-$10$} dataset with grid search, and apply the same parameters to all the remaining data partitions, including the extreme large-scale edge case setting. The detailed hyperparameters of different models for \textsc{Retina} and \textsc{CIFAR-$10$} are shown in Table~\ref{Retina_cifar_params}.

\textbf{FL hyperparameters:} For \textsc{Retina} and \textsc{CIFAR-$10$}, we set the number of local training epochs $E$ on each client to $1$ (unless otherwise stated) and the total number of communication rounds to $100$, with all local clients participating in FL training in each round. $\beta$ is selected from $\{0.1, 0.3, 0.5, 0.7, 0.9, 0.97, 0.99, 0.997\}$ for FedAVGM~\cite{hsu2019measuring}, and is set to $0.5$ and $0.3$ for Retina and CIFAR-$10$ dataset, respectively. In FedProx~\cite{li2018federated}, $\mu$ is set to $0.001$ for Retina dataset and $0.1$ for \textsc{CIFAR-$10$} dataset by selecting from $\{0.001, 0.01, 0.1, 1\}$.

For the CelebA dataset, we randomly sample $10$ clients in each round of FL learning for parallel FL methods. We set $E$ to $1$, the maximum train round to $30$ for CWT, and $1000$ for all the other parallel FL methods, to ensure each local client joins in FL training for around $30$ rounds. $\mu$ of FedProx is set to $0.001$ for CelebA dataset. We allow each client to share $5\%$ percentage of their data among each other for FedAVG-Share on all the compared datasets. The detailed hyperparameters are shown in Table~\ref{Retina_cifar_params} and Table~\ref{CelebA_params}. Please refer to our project page for an implementation to reproduce our results.

\begin{table*}[t]
\fontsize{8.5}{11}\selectfont
\centering
\caption{Table of hyperparameters for experiments on \textsc{Retina} and \textsc{CIFAR-$10$} with  ResNets~\cite{he2016deep}, EfficientNets~\cite{tan2019efficientnet}, ViTs and Swins~\cite{liu2021swin}. Gradient clip at global norm $1$ are applied to all models to stabilize the training. The learning rate is halved every $30$ epochs in the step decay scheduler.}
\setlength\tabcolsep{3.5pt}
\begin{tabular}{l | c  c c c c c c}
\Xhline{3\arrayrulewidth}
Models & Dataset & Split type & Total Round & Optimizer type & Warm-steps & LR decay & Base LR \\
\Xhline{0.5\arrayrulewidth}
ResNets-CWT & Retina \& CIFAR-10 & All    & 100 & SGD & 500 & cosine & 0.03  \\
EfficientNets-CWT & Retina & All & 100 & AdamW & 500 & cosine & 0.0005  \\
EfficietNets-CWT & CIFAR-10 & All & 100 & SGD & 500 & cosine & 0.03  \\
ViTs-CWT & Retina \& CIFAR-10 & All  & 100 & SGD & 500 & cosine & 0.003  \\
Swins-CWT & Retina \& CIFAR-10 & All & 100 & AdamW & 500 & cosine & $3.125\times10^{-5}$  \\

 \Xhline{0.5\arrayrulewidth}
ResNets-FedAVG & Retina \& CIFAR-10 & All & 100 & SGD & 500 & cosine & 0.03  \\
 EfficientNets-FedAVG & Retina & All & 100 & AdamW & 500 & cosine & 0.0005  \\
 EfficientNets-FedAVG & CIFAR-10 & All & 100 & SGD & 500 & cosine & 0.03  \\
ViTs-FedAVG & Retina \& CIFAR-10 & All & 100 & SGD & 500 & cosine & 0.03  \\
Swins-FedAVG & Retina \& CIFAR-10 & All & 100 & AdamW & 500 & cosine & $3.125\times10^{-5}$  \\

 \Xhline{0.5\arrayrulewidth}
ResNet(50)-FedAVGM~\cite{hsu2019measuring} & Retina & All & 100 & SGD & 0 & step & 0.03  \\
ResNet(50)-FedAVGM~\cite{hsu2019measuring}  & CIFAR-$10$ & All & 100 & SGD & 500 & cosine & 0.03  \\
ResNet(50)-FedProx~\cite{li2018federated} & Retina & All & 100 & SGD & 0 & step & 0.03  \\
ResNet(50)-FedProx~\cite{li2018federated}  & CIFAR-$10$ & All & 100 & SGD & 500 & cosine & 0.03  \\
ResNet(50)-FedAVG-Share~\cite{zhang2020personalized} &  Retina \&  CIFAR-$10$ & All & 100 & SGD & 500 & cosine & 0.03  \\

\Xhline{3\arrayrulewidth}
\end{tabular}
\vspace{-0mm}
\label{Retina_cifar_params}
\end{table*}

\begin{table*}[t]
\fontsize{8.5}{11}\selectfont
\centering
\caption{Table of hyperparameters for experiments on CelebA and OpenImage. All methods are optimized with SGD (momentum $0.9$ and no weight decay), and gradient clip at global norm $1$.}
\setlength\tabcolsep{3.5pt}
\begin{tabular}{l | c  c c c c c}
\Xhline{3\arrayrulewidth}
Models &  Avg. Total Round & Warm-steps & Optimizer type & LR decay & Base LR   \\
\Xhline{0.5\arrayrulewidth}
ResNet(50)-CWT &   30 & 500 &SGD & cosine & 0.03  \\
ResNet(50)-FedAVG &   30 & 500 &SGD & cosine & 0.03  \\
ResNet(50)-FedProx &  30 & 500 &SGD & cosine & 0.03  \\
ResNet(50)-FedAVG-Share &   30 & 500 &SGD & cosine & 0.03  \\
ViT(S)-CWT &   30 & 500 &SGD & cosine & 0.003  \\
ViT(S)-FedAVG &   30 & 500 &SGD & cosine & 0.03  \\
\Xhline{3\arrayrulewidth}
\end{tabular}
\vspace{-0mm}
\label{CelebA_params}
\end{table*}

\begin{figure*}[t]
	\centering
	\begin{center}
		\begin{tabular}{ccc}
\includegraphics[width=0.493\linewidth]{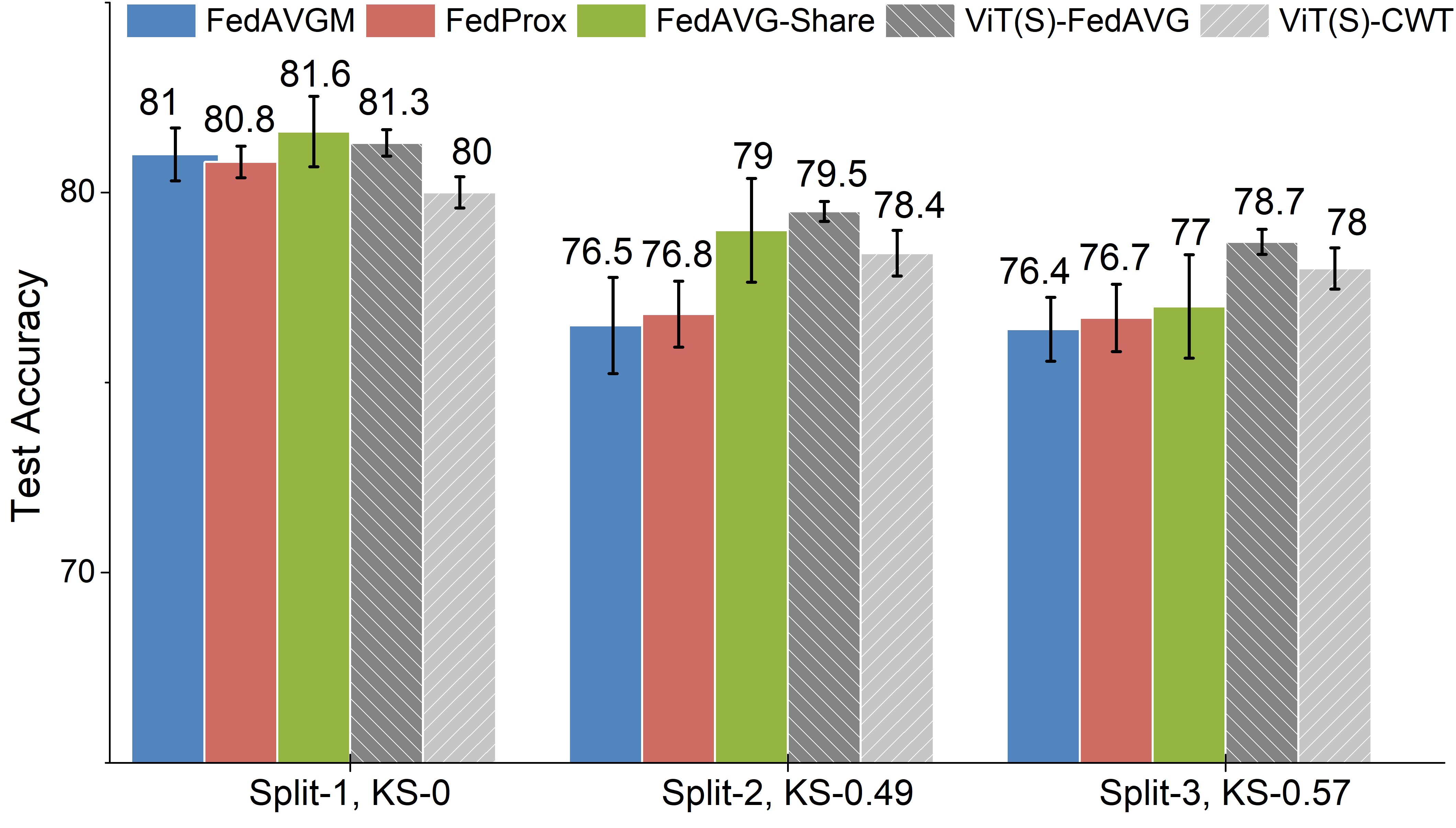} &  \hspace{-3mm}
	\includegraphics[width=0.5\linewidth]{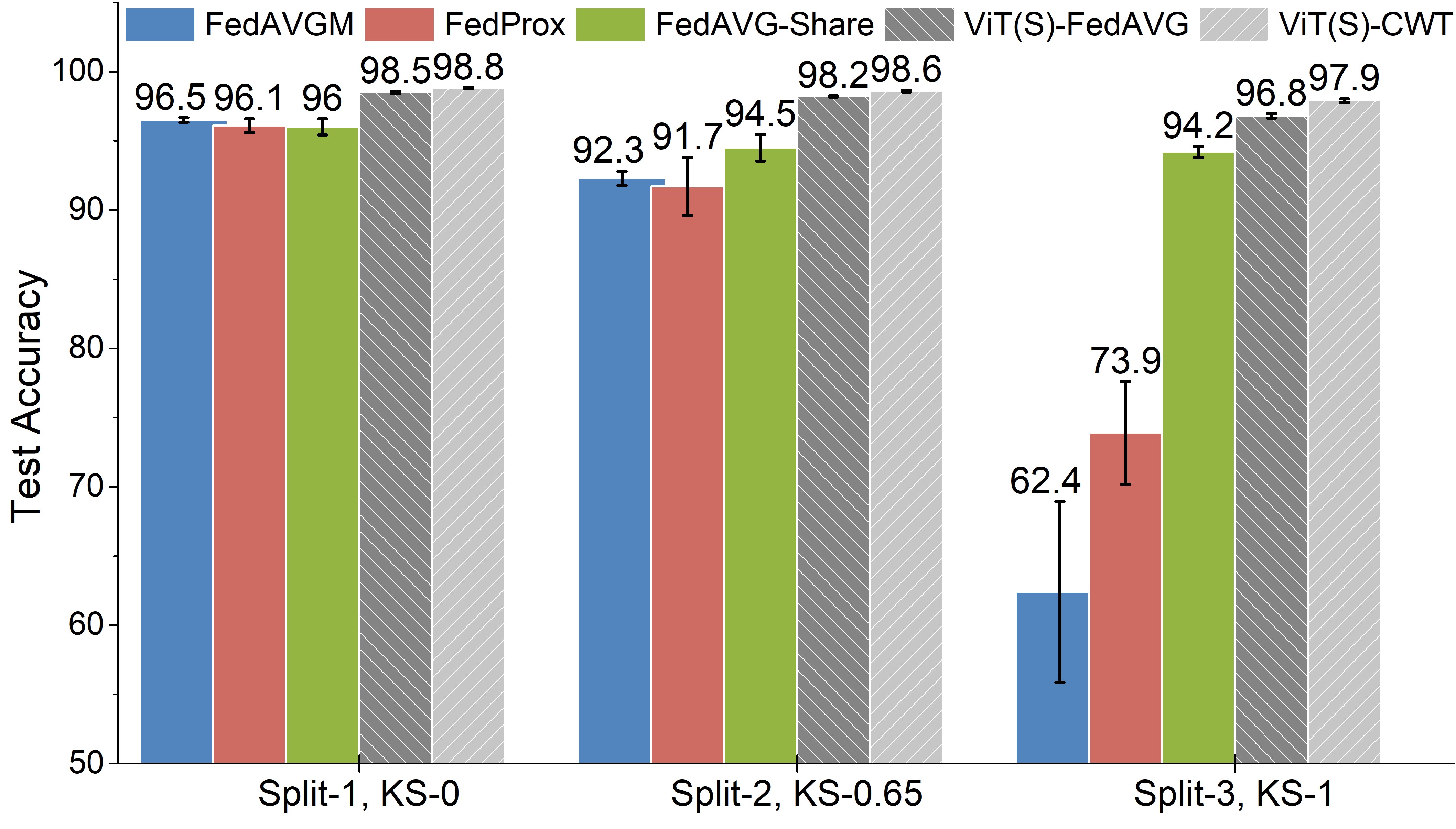} \\
 Retina dataset & CIFAR-$10$ dataset\\
		\end{tabular}
	\end{center}
\vspace{-3mm}
	\caption{Comparisons with state-of-the-art optimization based federated learning methods with ResNet-$50$ as backbone. Vision Transformer-based FL methods (ViT(S)-CWT and ViT(S)-FedAVG) outperform other methods in non-IID data partitions.}
	\vspace{-3mm}
	\label{fig:supple_comparisons}
\end{figure*}

\section{Additional Results}
\label{appendix:results}


\vspace{-1mm}
\subsection{Take-aways for Practical Usage}
\label{appendix:tips}
\vspace{-1mm}

The training strategy of \textsc{ViT} in FL can be directly inherited from \textsc{ViT} training, such as using linear warm-up and learning rate decay, and gradient clip. We also notice that gradient clip stabilizes training for most FL methods on the highly heterogeneous data partition, and therefore can be applied as a general technique in FL applications (see Figure~\ref{fig:gradient_clip} of ViT(B)-FL and ResNet(50)-FedAVG with and without gradient clip). The training of \textsc{ViT}-CWT favors a relatively smaller learning rate on heterogeneous data partitions, whereas using a smaller learning rate for CNN counterparts leads to worse performance. In real-world applications, users can use a large learning rate for IID or mild-skewed data partitions for \textsc{ViT}-CWT, but a smaller learning rate is necessary to stabilize training for highly heterogeneous data partitions.

\begin{table*}[t]
  \renewcommand{\arraystretch}{1}
  \centering
  \footnotesize
  \caption{Prediction accuracy (\%) of CWT and FedAVG on \textsc{Retina} and \textsc{CIFAR-$10$} when using ResNet50 and ResNet50(GN) as the backbone network. Replacing the batch normalization layer with group normalization in ResNet50 still suffers performance loss on highly heterogeneous data partitions, indicating that the promising performance of \names\ does not come purely from not using batch normalization.}
  \label{table:acc_CIFAR-$10$_hybrid}
  \begin{tabular}{c|ccc|ccc}
    \Xhline{3\arrayrulewidth}
&\multicolumn{3}{c|}{\textsc{Retina}}  &\multicolumn{3}{c}{\textsc{CIFAR-$10$}} \\
    & Split-1 & Split-2 & Split-3 & Split-1 & Split-2 & Split-3 \\
    \Xhline{0.5\arrayrulewidth}
    ResNet(50)-CWT & 79.44 & 77.01 & 71.30 & 96.08 & 56.46 & 19.92   \\
    ResNet(50)(GN)-CWT& 82.21 & 81.13 & 77.05 & 95.10 & 93.87 & 87.70   \\
    ResNet(50)-FedAVG  & 80.48 &	76.36 & 75.99 & 96.51 & 93.14 & 59.68 \\
    ResNet(50)(GN)-FedAVG & 82.40 &	80.13 &	80.57 & 96.39 & 95.12 & 86.20 \\
    \Xhline{3\arrayrulewidth}
  \end{tabular}
\end{table*}

\begin{table}[h]
\setlength\tabcolsep{4.15pt}
  \renewcommand{\arraystretch}{1}
  \centering
  \footnotesize
  \renewcommand{\tabcolsep}{2pt}
 \caption{Prediction accuracy (\%) on a large-scale real world dataset OpenImage [Ref.A], covering 365 categories across 9,265 clients. {\textsc{ViT}}s significantly outperform their ResNet (R in Table) counterparts.\vspace{-3mm}}
 \label{table:OpenImage}
  \resizebox{\linewidth}{!}{
    \begin{tabular}{cc|cccc}
    \Xhline{3\arrayrulewidth}
    \centering

 R-CWT & ViT-CWT    &  R-FedAVG & R-FedProx & R-FedAVG-Share & ViT-FedAVG  \\
    \Xhline{0.5\arrayrulewidth}
 $41.62 $ & $\textbf{64.39} $ &  $50.92 $ & 51.39 & 55.34 &$\textbf{67.95} $ \\
    \Xhline{3\arrayrulewidth}
  \end{tabular}
  }
    \vspace{-3mm}
\end{table}

\vspace{-1mm}
\subsection{Experiments on Real-World Federated Datasets}
We further evaluate on a large-scale real-world dataset, OpenImage image classification~\cite{lai2021fedscale} collected from Flickr, containing 1.3M images spanning 600 categories across 14k clients. We select the categories with \#samples per class between 20 and 800 from the dataset, resulting in 81,088 images spanning 365 categories across 9,265 clients. We use similar training parameters to CelebA for OpenImage, i.e., we randomly sample 10 clients in each round of FL learning for parallel FL methods. We set E to 1, the maximum train round to 30 for CWT, and 27,000 for all the other parallel FL methods, to ensure each
local client joins in FL training for around 30 rounds. From Table~\ref{table:OpenImage}, \ViT\ significantly outperforms ResNets on this heterogeneous large-scale real-world data partition, even outperforming ideal centrally-hosted models (60.56\% for ResNet and 63.50\%  for \ViT\ on centrally-hosted dataset)

\subsection{Investigating the Influence of Normalization Technique in \names}
\vspace{-1mm}

The batch normalization layer has been shown to be one of the major factors that deteriorate the performance of federated learning methods on non-IID data partitions~\cite{hsieh2020non,gupta2021addressing}. Hsieh \emph{et al.} \cite{hsieh2020non} demonstrate that group normalization (or layer normalization) can avoid the skew-induced accuracy loss of batch normalization on non-IID data. This may raise the question: does the promising performance of \names\ come purely from not using a batch normalization layer? To answer this question, we compare \names\ with FL-ResNet50 (GN) by replacing all batch normalization layers in ResNet(50) with group normalization. As shown in Table~\ref{table:acc_CIFAR-$10$_hybrid}, group normalization indeed helps to obtain better performance for both CWT and FedAVG on mildly skewed data partitions than their batch normalization counterparts. For example, the performance on Split-2 of CIFAR-$10$ is improved from original $56.46\%$ to $93.87\%$. However, it still suffers performance loss on highly skewed data partitions. In contrast, \names\ consistently shows promising results on both mildly skewed and extremely highly skewed data partitions  (see Figure~\ref{fig:fig_acc_on_Cifar10_Retina} in main body paper for our results), indicating that the effectiveness \names\ does not arise purely from different normalization techniques.

\vspace{-1mm}
\subsection{Comparisons to Existing FL Methods}
\label{appendix:comparison_more}
\vspace{-1mm}

We compare \names\ to several state-of-the-art optimization based FL methods: FedAVGM~\cite{hsu2019measuring}, FedProx~\cite{li2018federated}, and FedAVG-Share~\cite{zhang2020personalized}.
We use ResNet-50 as the backbone network for all the compared FL methods. We tune the best parameters (including learning rate, momentum parameter $\beta$ for FedAVGM, and penalty constant $\mu$ in the proximal term of FedProx) on Split-2 dataset with grid search, and apply the same parameters to all the remaining data partitions. We allow each client to share $5\%$ percentage of their data among each other for FedAVG-Share.

As shown in Figure~\ref{fig:supple_comparisons}, \names\ outperforms all the other FL methods in non-IID data partitions. Both FedProx~\cite{li2018federated} and FedAVGM~\cite{hsu2019measuring} suffer severe performance drops on highly heterogeneous data partitions despite carefully tuned optimization parameters. Similarly, FedAVG-Share also suffers from performance drops on highly heterogeneous data partition Split-3 even when $5\%$ percentage of the local data is shared among all clients ($94.2\%$ of Split-3 on CIFAR-$10$ dataset compared to $96\%$ on Split-1). We conclude that simply using Transformers achieve superior performance than several recent methods designed for federated optimization, which often require careful tuning of optimization parameters.

\end{document}